\documentclass[sigconf]{acmart}

\usepackage{booktabs} 

\usepackage[linesnumbered,boxed,ruled]{algorithm2e}
\usepackage{enumitem}
\usepackage{dsfont}
\usepackage{graphicx, subfigure}
\usepackage{balance}

\setcopyright{rightsretained}

\copyrightyear{2018} 
\acmYear{2018} 
\setcopyright{acmcopyright}
\acmConference[KDD 2018]{24th ACM SIGKDD International Conference on Knowledge Discovery \& Data Mining}{August 19--23, 2018}{London, United Kingdom}
\acmBooktitle{KDD 2018: 24th ACM SIGKDD International Conference on Knowledge Discovery \& Data Mining, August 19--23, 2018, London, United Kingdom}
\acmPrice{15.00}
\acmDOI{10.1145/3219819.3219844}
\acmISBN{978-1-4503-5552-0/18/08} 

\begin{document}
\title{Distributed Collaborative Hashing and Its Applications in Ant Financial}

\author{Chaochao Chen}
\affiliation{
  \institution{Ant Financial Services Group}
  \city{Hangzhou, China, 310099} 
}
\email{chaochao.ccc@antfin.com}

\author{Ziqi Liu}
\affiliation{
  \institution{Ant Financial Services Group}
  \city{Hangzhou, China, 310099} 
}
\email{ziqiliu@antfin.com}

\author{Peilin Zhao}
\affiliation{
  \institution{Ant Financial Services Group}
  \city{Hangzhou, China, 310099} 
}
\email{peilin.zpl@antfin.com}

\author{Longfei Li}
\affiliation{
  \institution{Ant Financial Services Group}
  \city{Hangzhou, China, 310099} 
}
\email{longyao.llf@antfin.com}

\author{Jun Zhou}
\affiliation{
  \institution{Ant Financial Services Group}
  \city{Hangzhou, China, 310099} 
}
\email{jun.zhoujun@antfin.com}

\author{Xiaolong Li}
\affiliation{
  \institution{Ant Financial Services Group}
  \city{Hangzhou, China, 310099} 
}
\email{xl.li@antfin.com}

\begin{abstract}
Collaborative filtering, especially latent factor model, has been popularly used in personalized recommendation. 
Latent factor model aims to learn user and item latent factors from user-item historic behaviors. 
To apply it into real big data scenarios, efficiency becomes the first concern, including offline model training efficiency and online recommendation efficiency. 
In this paper, we propose a \underline{\textbf{D}}istributed \underline{\textbf{C}}ollaborative \underline{\textbf{H}}ashing (\textbf{DCH}) model which can significantly improve both efficiencies.
Specifically, we first propose a distributed learning framework, following the state-of-the-art \emph{parameter server} paradigm, to learn the offline collaborative model. 
Our model can be learnt efficiently by distributedly computing subgradients in minibatches on workers and updating model parameters on servers asynchronously. 
We then adopt hashing technique to speedup the online recommendation procedure. 
Recommendation can be quickly made through exploiting lookup hash tables. 
We conduct thorough experiments on two real large-scale datasets. 
The experimental results demonstrate that, comparing with the classic and state-of-the-art (distributed) latent factor models, DCH has comparable performance in terms of recommendation accuracy but has both fast convergence speed in offline model training procedure and realtime efficiency in online recommendation procedure.
Furthermore, the encouraging performance of DCH is also shown for several real-world applications in Ant Financial. 
\end{abstract}

%
%



\maketitle

\section{Introduction}	
With the rapid growth of E-commerce and other online applications, all kinds of information appear on the internet.
Meanwhile, more and more users search, buy, and even study through online platforms. 
Naturally, it becomes much more difficult for users to find their desired items (information), which is known as the information overload problem. 
Personalized recommendation is widely exploited in almost all the internet companies due to its great ability to solve the information overload problem \cite{sarwar2001item,das2007google}.  

\begin{figure*}[!htb]
\centering
\subfigure[Original data] { \includegraphics[width=4.5cm,height=3.5cm]{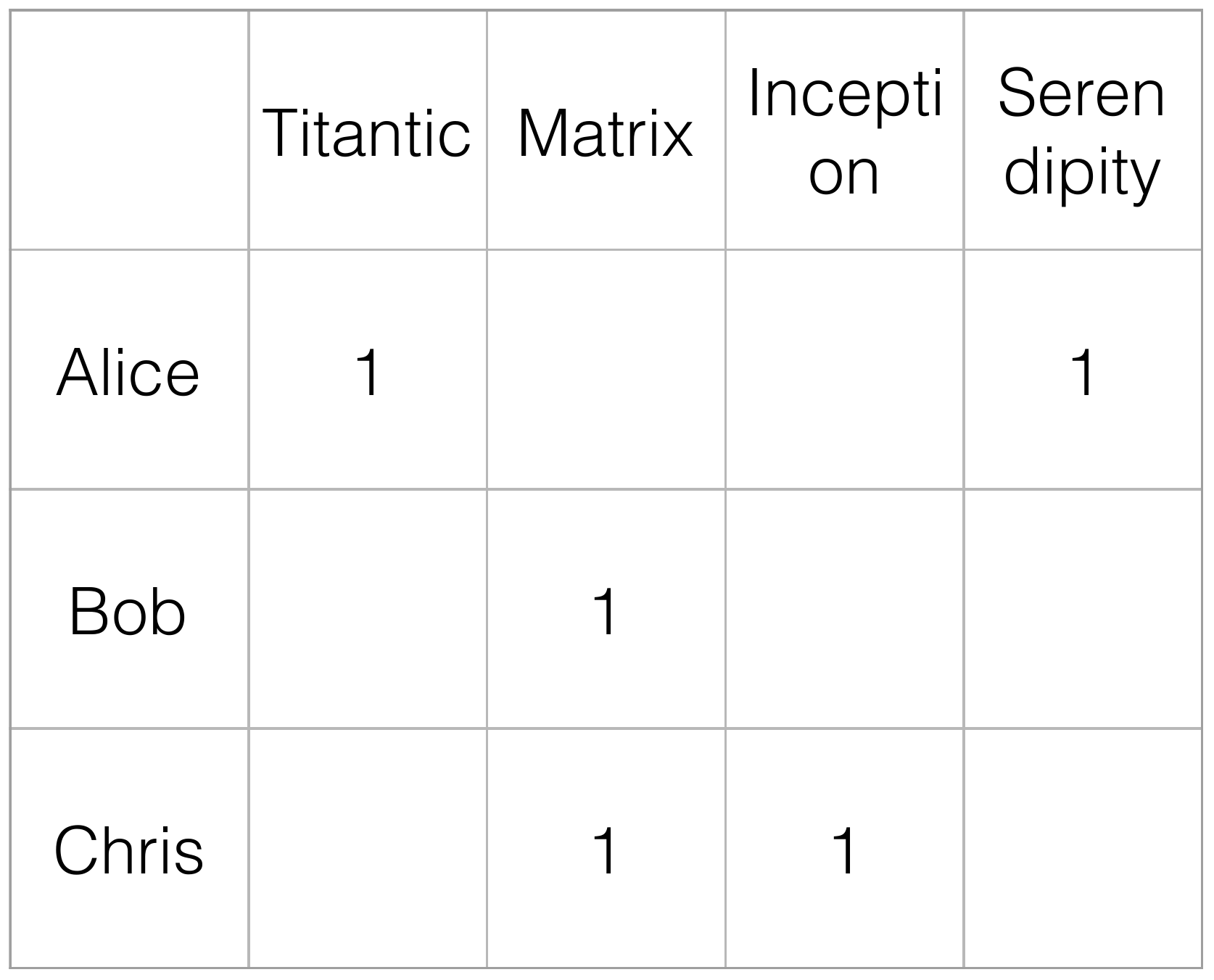}}~~~~~~~~~~~~~~~
\subfigure[Offline model training]{ \includegraphics[width=4.5cm,height=3.5cm]{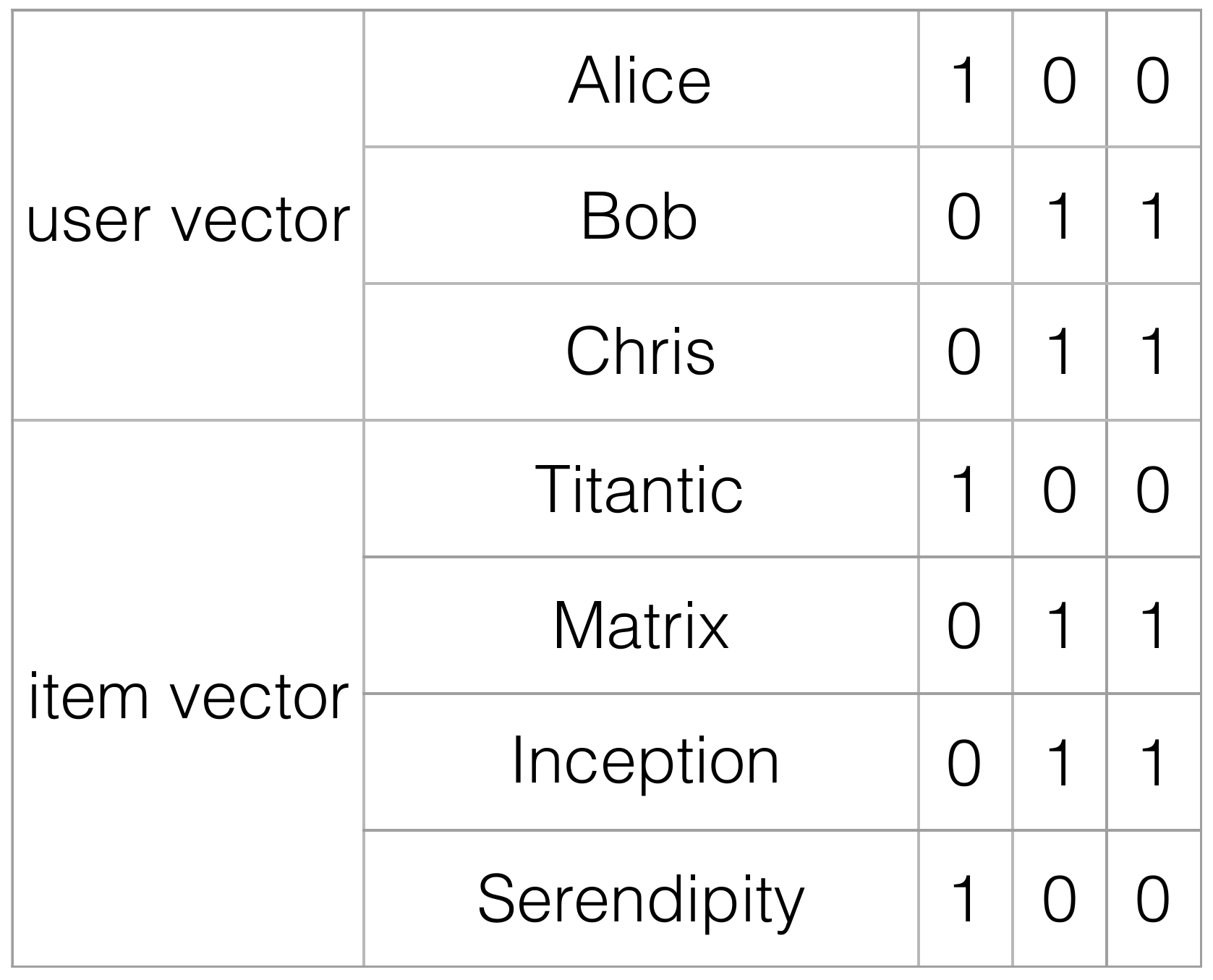}}~~~~~~~~~~~~~~~
\subfigure[Online recommendation]{ \includegraphics[width=4.5cm,height=3.5cm]{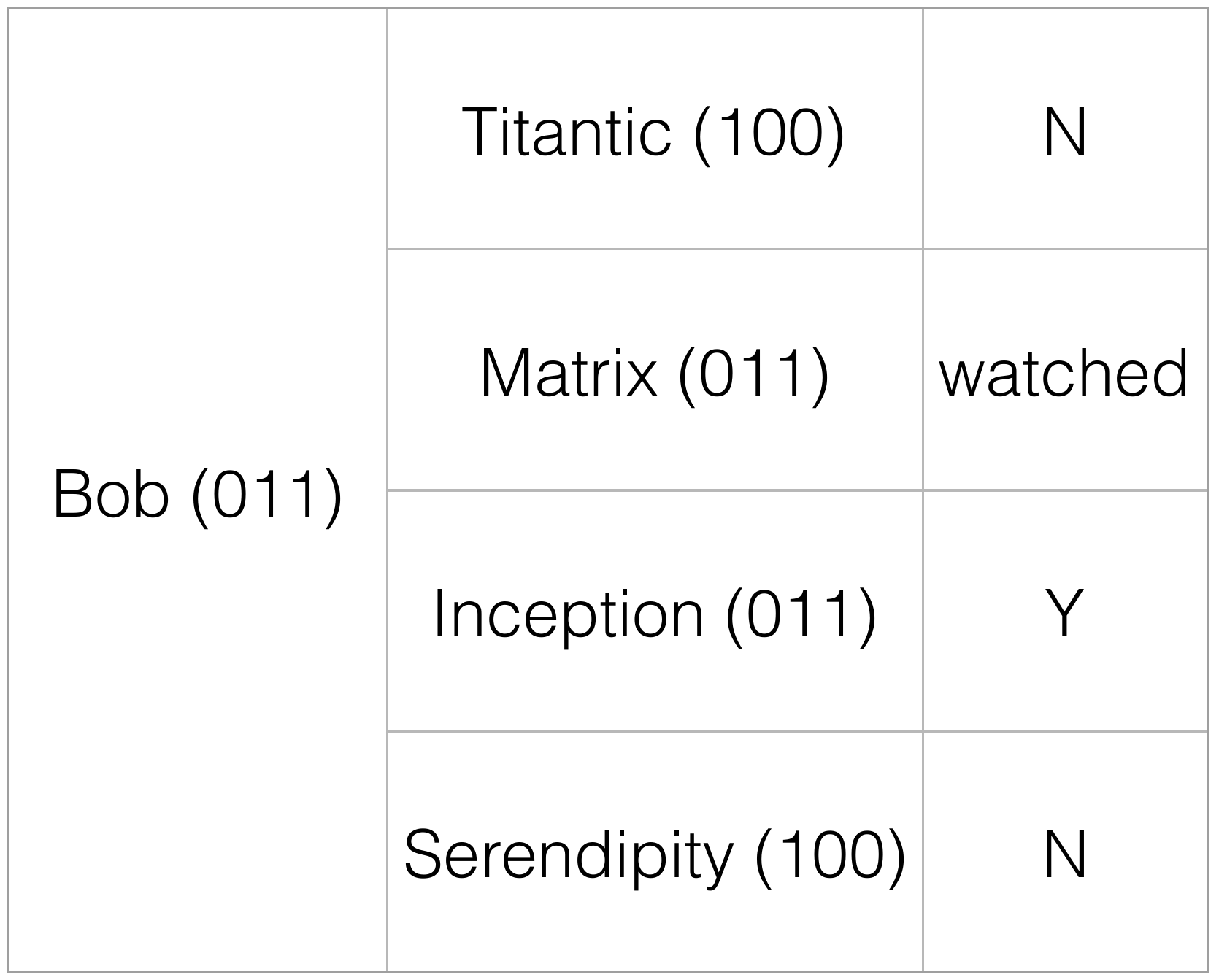}}
\caption{Toy exmaple. Using latent factor models to make recommendation includes two parts, i.e., offline model training and online recomendation.}
\label{toyexample}
\end{figure*}

Collaborative Filtering (CF) is one of the most popular techniques in personalized recommendation, and its main assumption is that users who behave similarly on some items will also behave similarly on other items \cite{resnick1994grouplens,su2009survey}. 
Among CF, model-based methods, especially latent factor models \cite{koren2008factorization,agarwal2009regression,rendle2010factorization}, draw a lot of attention, which mainly consist of two steps, i.e., offline model training procedure and online recommendation procedure. 
The \textbf{offline model training procedure} focuses on learning user and item latent factors from the known user-item actions, e.g., buy, rate, and click. 
Note that user and item latent factors are usually real-valued vectors with the same dimension and preserve the preferences of user and item respectively \cite{mnih2007probabilistic,koren2009matrix}.
The \textbf{online recommendation procedure} focuses on ranking the top $k$ items which have the most similar preferences with the target user. 
Take Figure \ref{toyexample} for example, during the offline model learning procedure, we learn that \emph{Bob} and \emph{Chris} have similar latent factors, since they both like scientific movies. 
Similarly, we get that \emph{Matrix} and \emph{Inception} have similar latent vectors, since they both belong to scientific movies. 
Here, we assume user and item latent factors are 3-dimensional binary vectors for simplification.
During the online recommendation procedure, we will naturally recommend \emph{Inception} to \emph{Bob} since their latent vectors are the same. 

To apply collaborative models in practical big data scenarios, efficiency becomes the first challenge, which includes both offline model training efficiency and online recommendation efficiency. 
On one hand, although most existing latent factor models scale linearly with the size of training dataset, a single machine has very limited memory and CPU.
Thus, it is still quite time-consuming to frequently update a collaborative model on a single machine due to the large user and item numbers in real applications. 
On the other hand, online recommendation needs to calculate the similarities of latent factors between the target user and all the candidate items.
This ranking procedure always needs to be done in realtime, and thus becomes a serious problem when the size of candidate items is very large. 

To solve the efficiency challenge of the existing collaborative models, in this paper, we propose a Distributed Collaborative Hashing 
(DCH) model. 
For offline model training, we develop a distributed framework following the state-of-the-art parameter server paradigm \cite{li2014scaling,schelter2014factorbird,li2016difacto}. 
Specifically, our model can be learnt efficiently by distributedly computing subgradients in minibatches on workers and updating model parameters on servers asynchronously. 
After that, our model outputs the hash codes as latent factors for each user and each item, which makes them easy to store, comparing with the real-valued latent vectors. 
For online recommendation, hashing techniques \cite{kulis2009kernelized,weiss2009spectral} are able to make efficient online recommendations with user and item hash codes.
This ranking procedure usually has linear and even constant time complexity by exploiting lookup hash tables \cite{wang2016learning} and thus satisfies the realtime requirement of online recommendation. 
We finally conduct thorough experiments on two large-scale datasets, i.e., the public \emph{Netflix} dataset and the real user-merchant payment data in Ant Financial.

Our main contributions are summarized as follows:
(1) We propose an efficient model named Distributed Collaborative Hashing (DCH), which offers the ability of efficient offline model training and online recommendation.
    To the best of our knowledge, this is the first attempt in literature to address the efficiency problems of both offline model training and online recommendation simultaneously.
(2) We develop a distributed learning algorithm, i.e., asynchronous stochastic gradient descent with bounded staleness, to optimize DCH using the state-of-the-art parameter server paradigm. 
(3) The experimental results on two large-scale real datasets demonstrate that, comparing with the classic and state-of-the-art (distributed) latent factor models, our model has both 
(a) comparable accuracy performance, and 
(b) fast convergence speed in offline model training procedure and realtime efficiency in online recommendation procedure.
(4) We have successfully deployed DCH into several real-world applications in Ant Financial and achieved encouraging performance.

\section{Background}\label{background}
In this section, we review background knowledge in three groups, i.e., (1) latent factor model, (2) hashing technique, and (3) parameter server.

\subsection{Latent Factor Model}
Latent factor model aims to learn user and item latent factors from the existing user-item action histories and other additional information, e.g., user social relationship, item content information, and contextual information \cite{mnih2007probabilistic,koren2008factorization,agarwal2009regression,rendle2010factorization,gemulla2011large,mcauley2013hidden,chen2014context,chen2016recommender}. 
Take a classic latent factor model, i.e., matrix factorization \cite{mnih2007probabilistic}, for example, it learns user and item latent factors through regression on the existing user-item ratings, which can be formalized as,
\begin{equation}
\label{lfm}
\begin{split}
\mathop {\arg \min }\limits_{u_i,v_j} \sum\limits_{i ,j} {{\left( {r_{ij} - u_i^T{v_j}} \right)}^2}  + \lambda\left(\sum\limits_{i} {||{u_i}||^2}  + \sum\limits_{j} {||{v_j}||^2}\right),
\end{split}
\end{equation}
where $u_i$ and $v_j$ denote the latent factors of user $i$ and item $j$, respectively, and $r_{ij}$ denotes the known rating of user $i$ on item $j$. We will describe other parameters in details later.

Most existing latent factor models aim to learn the real-valued user and item latent factors. 
Although they have good performance in terms of recommendation accuracy and scale well during offline training, it is difficult for them to rank the top $k$ neighbors efficiently during online recommendation. 
On one hand, it needs lots of space to store the real-valued user and item latent factors.
On the other hand, it is time-consuming to calculate user-item predictions and further rank scores to make recommendations.
Hashing technique provides an efficient way to solve this problem.

\subsection{Hashing Technique}
Hashing technique is aimed at learning binary hash codes of data entities and has been proven a promising approach to solve the nearest neighbor search problem \cite{kulis2009kernelized,salakhutdinov2009semantic,weiss2009spectral,liu2012supervised,norouzi2012fast}. 
It not only makes efficient in-memory storage of massive data feasible, but also makes the time complexity of the nearest neighbor search problem linear and even constant by exploiting lookup hash tables \cite{wang2016learning}.
Recently, there is a trend to adopt hashing techniques in personalized recommendation scenarios for better recommendation efficiency \cite{karatzoglou2010collaborative,zhou2012learning,ou2013comparing,zhang2014preference,smirnov2015locality,zhang2016discrete,lian2017discrete}.
They are mainly divided into two categories:
(1) first relax the solution from binary space $\{-1, 1\}$ to real space $[-1,1]$, and then learn real-valued user and item latent factors by using the same way as the existing latent factor models, and finally round or rotate them back to binary hash codes \cite{zhou2012learning,zhang2014preference}; 
(2) directly learn user and item binary hash codes by alternatively optimizing subproblems \cite{zhang2016discrete,lian2017discrete}. 

Although existing hashing based approaches efficiently solve the online recommendation problem, it is difficult for them to scale to large datasets in practice, since they do not support distributed model learning. 
Parameter server appears to solve this problem.

\subsection{Parameter Server}

\begin{figure}[t]
\centering
\includegraphics[width=\columnwidth]{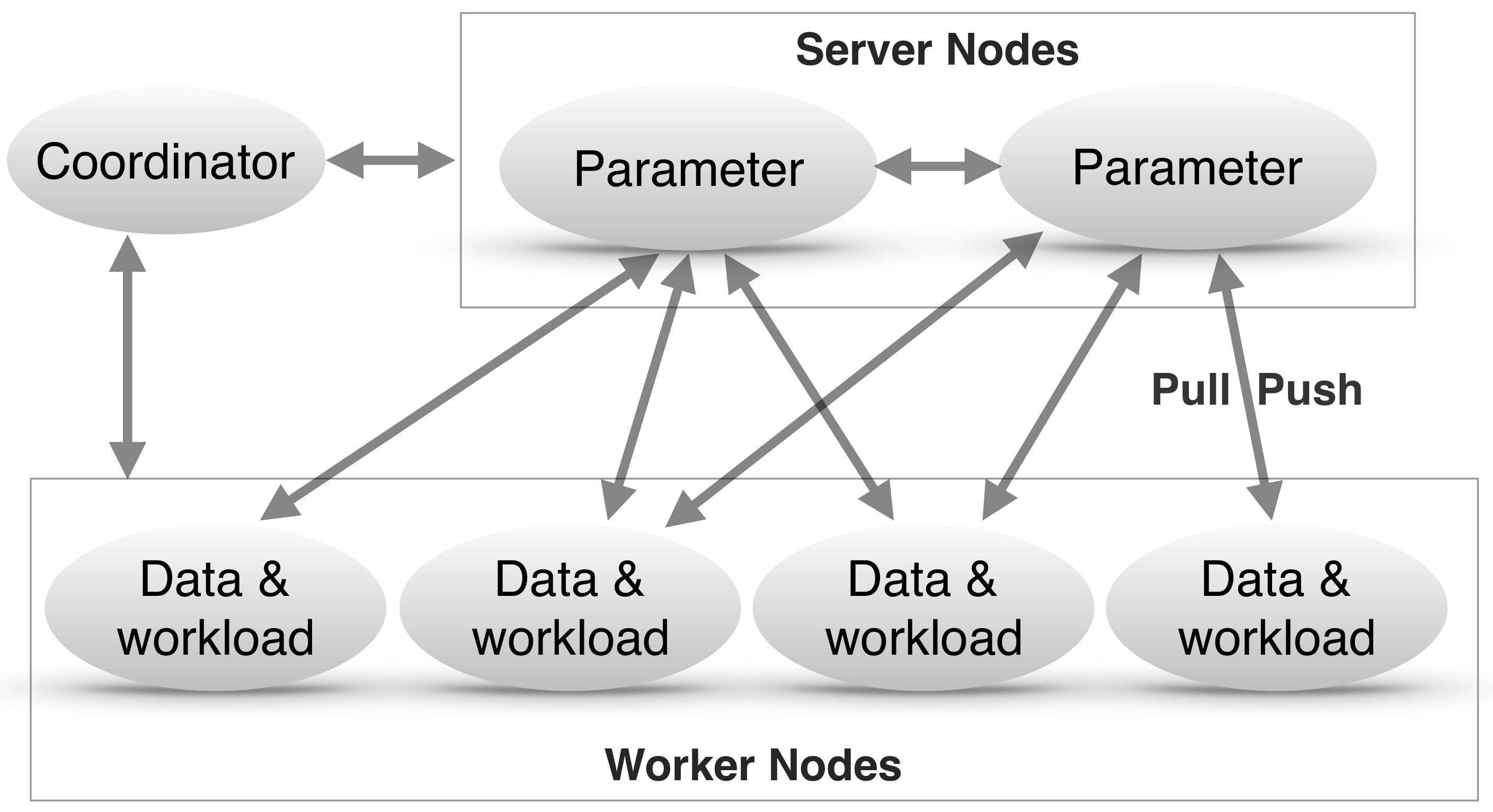}
\caption{Framework of parameter server, including machines layout and their communications.}
\label{ps}
\end{figure}

Parameter server is the state-of-the-art distributed learning framework \cite{li2014scaling,schelter2014factorbird,li2016difacto}.
It contains a coordinator and two other groups of computers, i.e., servers and workers, as is shown in Figure \ref{ps}. 
The coordinator controls the start and terminal of the algorithm based on a certain condition, e.g., the number of iterations. 
The servers are in charge of storing and updating model parameters of an algorithm, e.g., user and item latent factors of matrix factorization. 
The workers sequentially (or randomly) load and process training data and calculate the changes of the model parameters, e.g., the gradients of user and item latent factors when using gradient descent to optimize matrix factorization. 
Meanwhile, communications between workers and servers make sure the model parameters are correctly updated.
Specifically, communication is performed using the following two operations,
(1) \textbf{Pull(key, value)}. Before loading data, workers first pull the up-to-date model parameters from servers, and individually update these parameters. Note that, before pull operation, workers first check which parameters will be updated in this minibatch, and then only pull these parameters from servers. This sparse communication strategy greatly reduces the communication time, especially when model parameters are extremely large.
(2) \textbf{Push(key, value)}. After workers load data and calculate the changes of parameters, they push the changes of parameters to servers, and then let servers update these parameters.

The coordinator allows workers and servers to asynchronously update model parameters with bounded staleness, and thus is able to make fully use of the memory and CPU of each machine. 

\section{Distributed Collaborative Hashing}	
In this section, we first formulate our problem.
We then describe the collaborative hashing model that aims to learn user and item hash codes.
Next, we propose the distributed optimization method to learn the collaborative hashing model.
Then, we present our distributed implementation in details. 
Finally, we describe online recommendation and end this section with a discussion.

\subsection{Problem Formulation}
Collaborative Filtering (CF) solves the information overload problem through recommending latent interesting items to target users. 
To do this, CF first learns the real-valued latent factors of users and items from the past user-item action histories, and then makes recommendation through matching users' and items' latent factors. 
In contrast, collaborative hashing aims to learn the binary-valued latent factors of users and items. 
Formally, let $\mathcal{U}$ and $\mathcal{V}$ be the user and item set with $M$ and $N$ denoting user size and item size, respectively.
Let $r_{ij}$ be the known rating of user $i \in \mathcal{U}$ on item $j \in \mathcal{V}$. 
Usually, $r_{ij}$ is a real value in a certain region, and without loss of generality, we assume $r_{ij}$ ranges in [0, 1] in this paper.
Let $\mathcal{O}$ be the training dataset, where all the user-item ratings in it are known. 
We also use $\mathcal{O}_u$ and $\mathcal{O}_v$ to denote the users and items in the training dataset, respectively.
We further let $\textbf{U}\in \{-1, 1\}^{K\times{}M}$ and $\textbf{V}\in \{-1, 1\}^{K\times{}N}$  be the user and item latent feature matrices, with their column vectors $u_i$ and $v_j$ be the $K$-dimensional binary hash codes for user $i$ and item $j$, respectively. That is, $u_i \in \{-1, 1\} ^ K$ and $v_j \in \{-1, 1\} ^ K$, where $K$ is the length of the hash codes. 

Collaborative hashing aims to learn hash codes for each user $i \in \mathcal{U}$ and each item $j \in \mathcal{V}$, and then makes recommendation through matching the hash codes of users and items.

\subsection{Collaborative Hashing}

To make recommendation using the learnt hash codes, user and item hash codes need to preserve the preferences between them, i.e., the similarity between their hash codes should directly denote the similarity of their preferences. 
For two hash codes, e.g., $u_i$ and $v_j$, their similarity can be estimated by their Hamming distance, i.e., the common bits in $u_i$ and $v_j$, that is, 
\begin{equation}
\label{hamming}
\begin{split}
	sim(u_i, v_j) & = \frac{1}{K}\sum\limits_{k = 1}^K{\mathds{1}\left(u_i^{(k)}=v_j^{(k)}\right)} = \frac{1}{2} + \frac{1}{2K} u_i^T v_j,
\end{split}	
\end{equation} 
where $\mathds{1}(\cdot)$ is the indicator function that equals to 1 if the expression in it is true and 0 otherwise. 
To learn the approximate solutions of $u_i$ and $v_j$, we introduce the following objective function,
\begin{equation}
\label{obj}
\begin{split}
	\mathop {\arg \min }\limits_{u_i,v_j \in \{-1,1\}^K} \mathcal{L} = & \sum\limits_{i \in \mathcal{O}_u,j \in \mathcal{O}_v}{\left(r_{ij} - \frac{1}{2} - \frac{1}{2K} u_i^T v_j\right)}^2 \\
	& + \lambda \left( ||{ \sum\limits_{i \in \mathcal{O}_u} u_i }||_F^2  +  ||{ \sum\limits_{j \in \mathcal{O}_v} v_j }||_F^2 \right),
\end{split}	
\end{equation} 
where $\lambda \geq 0$ and $||\cdot||_F^2$ denotes the Frobenius norm \cite{zhou2012learning}.

Equation (\ref{obj}) has an intuitive explanation. 
Its first term constrains that a user-item rating is proportional to their similarity, i.e., the bigger a user-item rating is, the more similar their hash codes are. 
The second term requires the binary codes to be balanced, i.e., they have the equal chance to be -1 or 1, and $\lambda$ controls the balanced degree. 
The balance constraint is equivalent to maximizing the entropy of each bit of the binary codes, which indicates that each bit carries as much information as possible \cite{zhou2012learning}.

Relaxation is widely adopted to find the approximate solutions of binary values \cite{wang2013learning}. 
We first relax the solution space from $\{-1,1\}^K$ to $[-1,1]^K$, and then apply the continuous optimization techniques, e.g., stochastic gradient descent or coordinate descent, to solve Eq.(\ref{obj}), and at last, round the learnt real-valued solutions into $\{-1,1\}^K$.

Given the relaxed problem, we take the gradient of $\mathcal{L}$ with respect to $u_i$ and $v_j$ and get,
\begin{equation}
\label{gradient}
\begin{split}
\frac{\partial \mathcal{L}}{\partial u_i} = -\frac{1}{K} \sum\limits_{j \in \mathcal{O}_v}{\left(r_{ij} - \frac{1}{2} - \frac{1}{2K} u_i^T v_j\right)}v_j + 2\lambda\sum\limits_{i' \in \mathcal{O}_u}{u_{i'}},\\
\frac{\partial \mathcal{L}}{\partial v_j} = -\frac{1}{K} \sum\limits_{i \in \mathcal{O}_u}{\left(r_{ij} - \frac{1}{2} - \frac{1}{2K} u_i^T v_j\right)}u_i + 2\lambda\sum\limits_{j' \in \mathcal{O}_v}{v_{j'}}.
\end{split}
\end{equation} 

In this paper, we choose Stochastic Gradient Descent (SGD) to optimize our model.
Suppose $\alpha$ is the learning rate, $u_i$ and $v_j$ are updated as,
\begin{equation}
\label{update}
\begin{split}
u_i \leftarrow u_i - \alpha \frac{\partial \mathcal{L}}{\partial u_i},\\
v_j \leftarrow v_j - \alpha \frac{\partial \mathcal{L}}{\partial v_j}.
\end{split}
\end{equation} 

\textbf{Obtaining Binary Codes.}
After we solve the approximate solution of $\textbf{U}$ and $\textbf{V}$, we have the real-valued user and item latent vectors, i.e., $u_i$ and $v_j$ $\in {[-1,1]}^K$. 
Thus, we need to round the real-valued user and item latent factors into binary codes, i.e., $\widetilde{u}_i$ and $\widetilde{v}_j$ $\in {\{-1,1\}}^K$.
Following the work in \cite{zhou2012learning}, we round real-valued $u_i$ and $v_j$ to their closest binary vectors $\widetilde{u}_i$ and $\widetilde{v}_j$, so that the learnt binary codes are balanced. 
That is,
\begin{equation}
\label{roundu}
\begin{split}
{\widetilde{u}_i}^{(k)} = 
\begin{cases}
1,& \text{${u_i}^{(k)}$ $>$ median (${u_i}^{(k)}: i \in \mathcal{U}$)},\\
-1,& \text{otherwise},
\end{cases}\\
{\widetilde{v}_j}^{(k)} = 
\begin{cases}
1,& \text{${v_j}^{(k)}$ $>$ median (${v_j}^{(k)}: j \in \mathcal{V}$)},\\
-1,& \text{otherwise},
\end{cases}
\end{split}
\end{equation} 
where $k$ denotes the $k$-th bit in the binary codes and median($\cdot$) represents the median of a set of real numbers.

\subsection{Distributed Optimization---Asynchronous Stochastic Gradient Descent with Bounded Staleness}\label{boundedstaleness}

To fully make use of the memory and CPU of all the workers and servers, we adopt the idea of asynchronous stochastic gradient descent with bounded staleness to optimize our model. 
``Asynchronous'' means that all the workers independently run SGD on minibatch datasets. 
Specifically, each worker iteratively performs \textbf{a SGD operation}, i.e., it pulls model parameters (i.e., user and item latent factors) from servers, loads minibatch training data, calculates gradients, and pushes the gradients (i.e., model changes) to servers. 
Meanwhile, each server updates the model parameters stored in it once it receives the model gradients from workers. 
Since there are multi-worker running the SGD procedures asynchronously, each worker will probably not send the most recent model gradients to servers. 
The idea of ``bounded staleness (delay)'' has been extensively used in distributed frameworks \cite{li2014scaling}.
It makes a compromise between total synchronization and asynchronization, which means that servers will wait for all the workers to synchronize once after a certain period. 
We use parameter $P$ to denote the period of synchronization, i.e., synchronize once after each worker has done $P$ SGD procedures.

\begin{figure}[!tp]
\centering
\includegraphics[width=7cm]{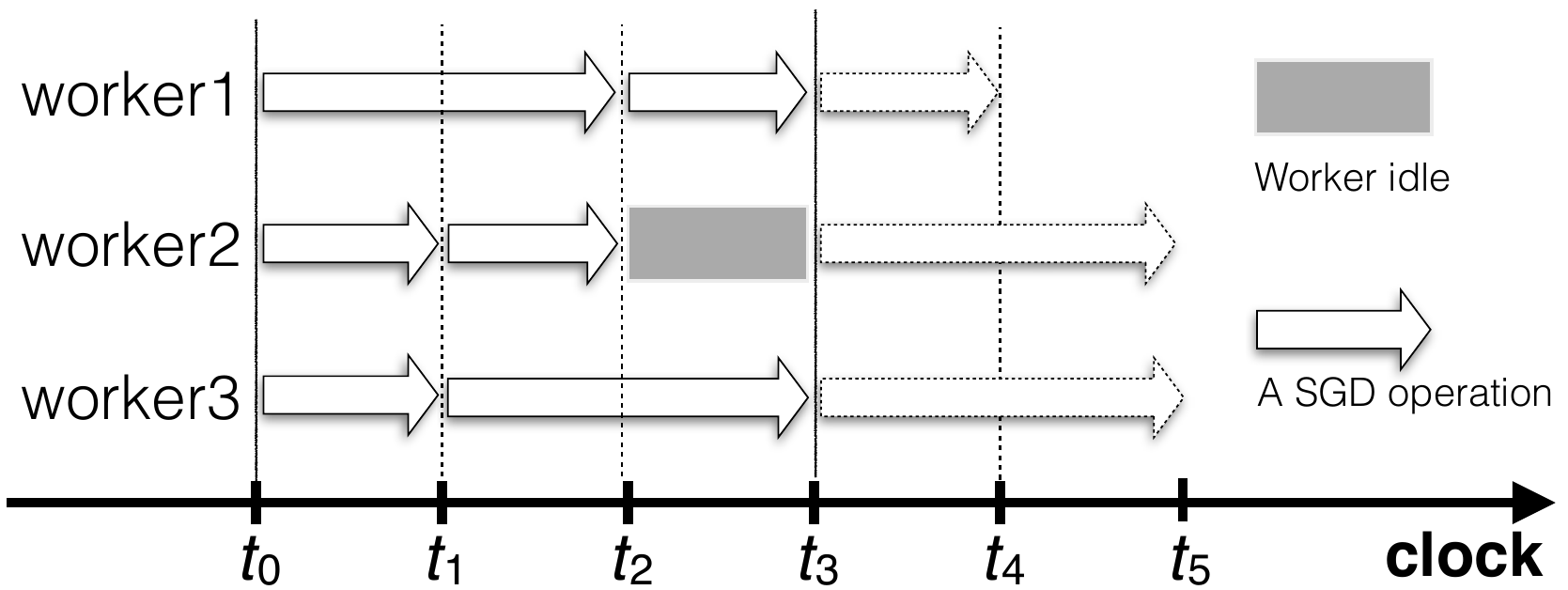}
\caption{An illustration of asynchronous SGD with bounded staleness ($P=2$).}
\label{nsync}
\end{figure}

Figure \ref{nsync} shows the illustration of an asynchronous SGD with bounded staleness when $P=2$, where we omit the time of server operations for conciseness. 
Three workers, i.e., $W_1$, $W_2$, and $W_3$, start to perform SGD operations at $t_0$. 
$W_1$ takes three time slides (from $t_0$ to $t_3$) to finish two SGD iterations---two time slides (from $t_0$ to $t_2$) for the first iteration and one time slide (from $t_2$ to $t_3$) for the second iteration. 
Similarly, $W_2$ takes two time slides (from $t_0$ to $t_2$) to finish two SGD iterations, and $W_3$ takes three time slides (from $t_0$ to $t_3$) to finish two SGD iterations. 
During these two iterations, all the workers perform SGD operations independently.
However, servers will wait for all the workers to synchronize after the second iteration at time $t_3$, since we set $P=2$ in this example. 
After then, all the workers start to perform SGD operations independently again at $t_3$.
From it, we can see that synchronous SGD is a special case of asynchronous SGD with bounded staleness when $P=1$. 

\textbf{Model parameters projection.}
Model parameters may diverge when performing distributed SGD. 
To avoid this, gradient-projection approach becomes a choice to limit the set of admissible solutions in the ball of radius $1/\sqrt \gamma$ \cite{shalev2011pegasos}.
That is, we project $u_i$ and $v_j$ into this sphere by performing the following update at each synchronization period $t$, e.g., $t_3$ in Figure \ref{nsync}, that is
\begin{equation}
\label{projection}
\begin{split}
u_{i,t} \leftarrow min\left\{ 1, \frac{1/\sqrt \gamma}{||u_{i,t}||} \right\} u_{i,t},\\
v_{j,t} \leftarrow min\left\{ 1, \frac{1/\sqrt \gamma}{||v_{j,t}||} \right\} v_{j,t}.
\end{split}
\end{equation} 
where $u_{i,t}$ and $v_{j,t}$ denote latent vectors of $u_i$ and $v_j$ at time $t$, respectively.

\subsection{Implementation}
We implement DCH on Kunpeng---a parameter-server based distributed learning system in Alibaba and Ant Financial \cite{zhou2017kunpeng}. 
We summarize the implementation sketch in Algorithm 3.1, which contains the following three parts:


\begin{algorithm}[!htb]
\caption{Implement DCH on parameter server}\label{learning}
\KwIn {ratings in training set ($\mathcal{O}$), worker number ($W$), server number ($S$), synchronization period ($P$), minibatch size ($B$), learning rate ($\alpha$), projection radius ($\gamma$)}
\KwOut{user and item hash code matrices $\textbf{U}$ and $\textbf{V}$}
\underline{\textbf{Coordinator node:}}\\
\For{$s=1$ to $S$}{
	Assign server $s$ to do initialization 
}
Partition training data $\mathcal{O}$ evenly based on $W$ \\
\Repeat{convergence}
{
	\For{$p=1$ to $P$}{
		\For {$w=1$ to $W$, parallel}{
			Assign worker $w$ to do a SGD operation
		}
	}
	Wait all workers to synchronize \\
	\For{$s=1$ to $S$}{
		Assign server $s$ to do projection operation 
	}
}
\For{$s=1$ to $S$}{
	Assign server $s$ to do round operation 
}
\underline{\textbf{Worker node $w$:}}\\
\If{receive a SGD command from the coordinator}{
	Load a minibatch size ($B$) data \\
	Pull $u_i$ and $v_j$ that appear in this minibatch from server nodes \\
	Compute the gradients based on Eq.(\ref{gradient}) \\
	Push gradients back to servers
}
\underline{\textbf{Server node $s$:}}\\
\If{receive initialization command from the coordinator}{
	Initialize $\textbf{U}$ and $\textbf{V}$
}
\If{receive gradients from a worker}{
	Update $\textbf{U}$ and $\textbf{V}$ based on Eq.(\ref{update})
}
\If{receive projection command from the coordinator}{
	Project $\textbf{U}$ and $\textbf{V}$ based on Eq.(\ref{projection})
}
\If{receive round command from the coordinator}{
	Round $\textbf{U}$ and $\textbf{V}$ based on Eq.(\ref{roundu})
}
\Return $\textbf{U}$ and $\textbf{V}$
\end{algorithm}

\textbf{Coordinator node} controls the status and procedure of the whole algorithm, including starts and terminates of the algorithm, and gives commands to workers and servers.
Specifically, it first assigns servers to initialize $u$ and $v$, and then partitions the training data evenly based on $W$.
During the asynchronization period, it assigns workers to do SGD operations independently.
At each synchronization point, i.e., when the current iteration number $t\%p=0$, it waits all workers to synchronize and then assigns servers to do parameter projection. 
Finally, it assigns servers to do a round procedure after the model converges.

\textbf{Worker nodes} load data and perform computation.
Once they receive SGD commands from the coordinator, they first randomly load a minibatch data. 
They then pull the old user and item factors that will be used in the minibatch data from servers. 
After that, they calculate the gradients of user and item latent factors, and finally push them to servers.

\textbf{Server nodes} store and update $\textbf{U}$ and $\textbf{V}$. 
They initialize $\textbf{U}$ and $\textbf{V}$ at the beginning.
Then, once receive gradients from a worker, they update the corresponding user and item latent factors.
They will also project the solutions of $\textbf{U}$ and $\textbf{V}$ at each synchronization point.
Finally, they round the real-valued solutions to hash codes after the model converges.

\subsection{Online Recommendation}\label{online}
After user and item hash codes are learnt, online recommendation is extremely efficient. 
Hash code length (i.e., dimension of latent factors) and the number of candidate items are two factors that affect online recommendation efficiency. 
Hamming distance search \cite{ray2006efficient}, hashing lookup\cite{salakhutdinov2009semantic}, Hamming ranking \cite{liu2012supervised}, and multi-index hashing \cite{norouzi2012fast} are widely used for searching nearest neighbors in hashing techniques. 
Hamming distance search retrieves items with binary codes within a certain Hamming distance to the binary codes of the target user.
Hamming distance can be quickly calculated by using the \emph{XOR} operations between user and item hash codes, and thus is constant w.r.t the hash code length.
Hashing lookup returns items with binary codes within a certain Hamming distance to the binary codes of the target user from hash tables, and thus its time complexity 
is constant w.r.t the dataset size.
Hamming ranking ranks items according to their hamming distance with the target user and returns fixed items. 
Speedup can be achieved by fast hamming distance computation, i.e. popcnt function \cite{zhang2014preference}.
Multi-index hashing splits the original hashing code into several subcodes and conducts hashing table lookup seperately, which improves hashing lookup. 
We will further perform experiments in Section \ref{efficiency} to show the online recommendation efficiency of hashing techniques comparing with the existing latent factor models.

\subsection{Discussion}
Recently, graph representation has been drawing a lot of attention in both academic and industry, and can be applied into many tasks such as recommendation \cite{eksombatchai2017pixie}. 
Graph representation aims to learn latent representations of nodes on graphs, which can be seen as an improvement of the classic matrix factorization technique by considering complicated structural information on graph \cite{cao2015grarep}. 
Existing research has proven that most graph representation methods, e.g., DeepWalk \cite{perozzi2014deepwalk} and node2vec \cite{grover2016node2vec}, are in theory performing implicit matrix factorizations. 
Intuitively, these methods suffer from online recommendation efficiency problem when applying to recommendation tasks. 
Therefore, our proposed distributed collaborative hashing can be natually used to solve the efficiency problem. 
That is, we learn hash code representations instead of real-valued ones for nodes on graph, which is one of our future works.

\section{Experiments and Applications}	
In this section, we empirically compare the performance of the proposed DCH with the classic and state-of-the-art distributed collaborative models, including recommmendation accuracy performance and recommendation efficiency performance. 
We also study the parameters on its model performance and describe its the applications at Ant Financial.

\subsection{Setting}

\textbf{Datasets.} We use two large-scale datasets, i.e., \emph{Netflix} dataset and the real user-merchant payment data in \emph{Alipay}. 
\emph{Netflix} dataset is famous from the Netflix competition\footnote{The dataset is available at: https://www.kaggle.com/netflix-inc/netflix-prize-data/data} \cite{bennett2007netflix}, and is widely used due to its publicity and authority. 
\emph{Alipay} is a product of Ant Financial, which is also the biggest third-party online payment platform in China, and through it users are able to delivery both online and offline payment. 
Our \emph{Alipay} dataset consists of two parts.
The first part is sampled from user-merchant payment records during 2015/11/01 to 2017/10/31 whose ratings are taken as `1'.
The second part is sampled from the non-payment data whose ratings are taken as `0'.
Combining both parts, we get the \emph{Alipay} dataset.
We show their statistics in Table \ref{dataset}.

\textbf{Metrics.} We adopt two metrics to evaluate our model performance, i.e., precision and Discounted Cumulative Gain (DCG). 
Both of them are extensively used to evaluate the quality of rankings \cite{zhou2012learning,liu2014collaborative}. 
For each user, precision and DCG are defined as:
\begin{equation}\nonumber
\begin{split}
	&Precision@k=\frac{Number~of~postive~items~in~Top~k}{k},\\
	&DCG@k=\sum\limits_{i = 1}^k{\frac{2^{r_i}-1}{log(i+1)}},
  \end{split}
\end{equation} 
where $r_i$ denotes the rating of the $i$-th retrieved item. 
We take items whose ratings are equal to 5 as positive items on \emph{Netflix} dataset, and take each user-merchant payment as a positive data that a user rates a merchant (item) on \emph{Alipay} dataset. 

We split both datasets with two strategies: (1) randomly sample 80\% as training set and the rest 20\% as test set, and (2) randomly sample 90\% as training set and the rest 10\% as test set.
We use \emph{Netflix80} and \emph{Alipay80} to denote the first strategy, and use \emph{Netflix90} and \emph{Alipay90} to denote the second strategy.

\begin{table}
\centering
\caption{Dataset description}
\label{dataset}
\begin{tabular}{|c|c|c|c|}
  \hline
  Dataset & \#user & \#item & \#rating  \\
  \hline
  \hline
  \emph{Netflix} & 480,189 & 17,770 & 100,480,507 \\
  \hline
  \emph{Alipay} & 9,111,142 & 443,043 & 1,599,288,565 \\
  \hline
\end{tabular}
\end{table}

\textbf{Comparison methods.} 
We compare our proposed DCH with the following classic and state-of-the-art distributed models: 
\begin{itemize}[leftmargin=*] \setlength{\itemsep}{-\itemsep}
    \item \textbf{Matrix Factorization (MF)} is a classic collaborative model \cite{mnih2007probabilistic}, and it factorizes the user-item rating matrix into real-valued user and item latent factor matrixes without rounding them into hash codes.
    \item \textbf{MFH} first learns the real-valued user and item latent factors the same way as MF, and then rounds user and item latent factors into hash codes using Eq.(\ref{roundu}).
    \item \textbf{Distributed Factorization Machine (DFM)} is an implementation of the state-of-the-art DiFacto model \cite{li2016difacto} on parameter server.
    \item \textbf{DFMH} first learns the real-valued user and item latent factors the same way as DFM, and then rounds user and item latent factors into hash codes using Eq.(\ref{roundu}).
\end{itemize}

\textbf{Hyper-parameters.} 
We set worker number $W=50$, server number $S=20$, synchronization period $P=2$, learning rate $\alpha=0.001$, and minibatch size $B=1000$.
Besides, since our relaxed solution space is $[-1,1]$, we set projection radius $1/\sqrt{\gamma}=1$, i.e., $\gamma=1$. 
That is, we project the factors of users and items to 1 if they are bigger than 1 and to -1 if they are smaller than -1.
For the latent factor dimension $K$, we find its best value in $\{5,10,15,20,25,30\}$.
We find the best values of other hyper-parameters in $\{10^{-4},10^{-3},10^{-2},10^{-1},10^{0},$ $10^{1},10^{2}\}$.

\subsection{Comparison Results}
To verify our model performance, we compare DCH with four classic and state-of-the-art models on both \emph{Netflix} and \emph{Alipay} datasets. 
During the comparison, we use grid search to find the best parameters of each model. 
We report the comparison results on \emph{Netflix} in Table \ref{netflix5} and Table \ref{netflix10}, where $K=5$ and $K=10$ respectively, and report the comparison results on \emph{Alipay} in Table \ref{alipay5} and Table \ref{alipay10}, where $K=5$ and $K=10$ respectively.  
From them, we find that:
\begin{itemize}[leftmargin=*] \setlength{\itemsep}{-\itemsep}
    \item Recommendation performances of all the models increase with training data size. 
    For example, the Precision@5 of DCH increase 2.52\% on \emph{Netflix90} comparing with that on \emph{Netflix80} when $K=10$. 
    This is due to the data sparsity problem, i.e., recommendation accuracy decays when training data becomes sparser. 
    \item Recommendation performances of all the models are affected by the dimension of latent factor or hash code length. 
    Close observation shows the following results: recommendation accuracy of all the models are better when $K=10$ than $K=5$ on \emph{Netflix} datasets.
    On the contrary, some of them are better when $K=5$ than $K=10$ on \emph{Alipay} datasets, especially on \emph{Alipay80}.
    Note that the rating densities of \emph{Netflix} and \emph{Alipay} are 1.18\% and 0.04\%, respectively. 
    Different latent factor dimension or hash code length will contains different amount of information. 
    Small dimension/length will cause information loss, while large dimension/length may cause over-fitting.
    Since the rating density of \emph{Alipay} is much smaller than that of \emph{Netflix}, and \emph{Alipay80} has even sparser ratings, over-fitting is easier to appear on \emph{Alipay80} with the same $K$. 
    This is why the best value of $K$ on \emph{Alipay} is smaller than that on \emph{Netflix}.
    \item DCH achieves better results than the other two hash-based methods, i.e., MFH and DFMH, in almost all the cases.
    This is because, DCH directly optimizes the difference between rating and hamming distance, as is shown in Eq. (\ref{hamming}).
    In contrast, both MF and DFM optimize the difference between rating and the product of user-item's latent factors, as is shown in Eq. (\ref{lfm}).
    \item DCH has the \emph{comparable performance} with the classic and state-of-the-art distributed latent factor models, i.e., MF and DFM. 
    This is consistent with the existing research \cite{zhou2012learning}, i.e., hash technique can achieve comparable prediction accuracy performance with the classic latent factor models by using the appropriate values of $K$.
\end{itemize}

\begin{table*}[!htb]\small
\centering
\caption{Performance comparison on \emph{Netflix} datasets ($K=5$)}
\label{netflix5}
\begin{tabular}{|c|c|c|c|c|c|c|c|c|c|}
 \hline
  Datasets & \multicolumn{4}{c|}{\emph{Netflix80}} & \multicolumn{4}{c|}{\emph{Netflix90}}  \\
  \hline
  \hline
  Metrics & Precision@5 & Precision@10 & DCG@5 & DCG@10 & Precision@5 & Precision@10 & DCG@5 & DCG@10 \\
  \hline
  MFH & 0.2580 & 0.2473 & 44.9102 & 68.2807 & 0.2644 & 0.2573 & 45.2374 & 69.1247  \\
  \hline
  DFMH & 0.2572 & 0.2472 & 44.8933 & 68.2634 & 0.2641 & 0.2570 & 45.2339 & 69.1099  \\
  \hline
  MF & 0.2589 & 0.2479 & 44.9351 & 68.3471 & 0.2658 & 0.2584 & 45.3774 & 69.3064  \\
  \hline
  DFM & 0.2589 & 0.2481 & 45.0031 & 68.4398 & 0.2658 & 0.2586 & 45.3790 & 69.3621  \\
  \hline
  DCH & 0.2584 & 0.2478 & 44.9222 & 68.3279 & 0.2645 & 0.2576 & 45.2383 & 69.1353  \\
  \hline
\end{tabular}
\end{table*}

\begin{table*}[!htb]\small
\centering
\caption{Performance comparison on \emph{Netflix} datasets ($K=10$)}
\label{netflix10}
\begin{tabular}{|c|c|c|c|c|c|c|c|c|c|}
 \hline
  Datasets & \multicolumn{4}{c|}{\emph{Netflix80}} & \multicolumn{4}{c|}{\emph{Netflix90}}  \\
  \hline
  \hline
  Metrics & Precision@5 & Precision@10 & DCG@5 & DCG@10 & Precision@5 & Precision@10 & DCG@5 & DCG@10 \\
  \hline
  MFH & 0.2583 & 0.2478 & 44.9184 & 68.2782 & 0.2644 & 0.2572 & 45.2518 & 69.1101  \\
  \hline
  DFMH & 0.2579 & 0.2472 & 44.8933 & 68.2634 & 0.2644 & 0.2571 & 45.2484 & 69.1152  \\
  \hline
  MF & 0.2587 & 0.2479 & 44.9645 & 68.3807 & 0.2650 & 0.2576 & 45.2911 & 69.1707  \\
  \hline
  DFM & 0.2585 & 0.2482 & 44.9404 & 68.3803 & 0.2654 & 0.2576 & 45.3311 & 69.2036  \\
  \hline
  DCH & 0.2584 & 0.2479 & 44.9484 & 68.3808 & 0.2649 & 0.2574 & 45.2631 & 69.1472  \\
  \hline
\end{tabular}
\end{table*}

\begin{table*}[!htb]\small
\centering
\caption{Performance comparison on \emph{Alipay} datasets ($K=5$)}
\label{alipay5}
\begin{tabular}{|c|c|c|c|c|c|c|c|c|c|}
 \hline
  Datasets & \multicolumn{4}{c|}{\emph{Alipay80}} & \multicolumn{4}{c|}{\emph{Alipay90}}  \\
  \hline
  \hline
  Metrics & Precision@5 & Precision@10 & DCG@5 & DCG@10 & Precision@5 & Precision@10 & DCG@5 & DCG@10 \\
  \hline
  MFH & 0.200322 & 0.200266 & 0.590591 & 0.909867 & 0.200692 & 0.200304 & 0.591730 & 0.910686  \\
  \hline
  DFMH & 0.200443 & 0.200300 & 0.591013 & 0.910227 & 0.200714 & 0.200396 & 0.591818 & 0.910690  \\
  \hline
  MF & 0.200550 & 0.200319& 0.591969 & 0.911047 & 0.200992 & 0.200374 & 0.593726 & 0.912198  \\
  \hline
  DFM & 0.200565 & 0.200335 & 0.592235 & 0.911388 & 0.201059 & 0.200415 & 0.594371 & 0.912825  \\
  \hline
  DCH & 0.200329 & 0.200323 & 0.590738 & 0.910408 & 0.200814 & 0.200403 & 0.592797 & 0.911733  \\
  \hline
\end{tabular}
\end{table*}

\begin{table*}[!htb]\small
\centering
\caption{Performance comparison on \emph{Alipay} datasets ($K=10$)}
\label{alipay10}
\begin{tabular}{|c|c|c|c|c|c|c|c|c|c|}
 \hline
  Datasets & \multicolumn{4}{c|}{\emph{Alipay80}} & \multicolumn{4}{c|}{\emph{Alipay90}}  \\
  \hline
  \hline
  Metrics & Precision@5 & Precision@10 & DCG@5 & DCG@10 & Precision@5 & Precision@10 & DCG@5 & DCG@10 \\
  \hline
  MFH & 0.200326 & 0.200274 & 0.590664 & 0.909959 & 0.200809 & 0.200432 & 0.592072 & 0.910968  \\
  \hline
  DFMH & 0.200345 & 0.200289 & 0.590554 & 0.909863 & 0.200671 & 0.200422 & 0.591603 & 0.910554  \\
  \hline
  MF & 0.200394 & 0.200297 & 0.590823 & 0.910196 & 0.200839 & 0.200435 & 0.593061 & 0.911767  \\
  \hline
  DFM & 0.200416 & 0.200328 & 0.591269 & 0.910491 & 0.200921 & 0.200438 & 0.593078 & 0.911937  \\
  \hline
  DCH & 0.200356 & 0.200295 & 0.590731& 0.909967 & 0.200821 & 0.200436 & 0.592246 & 0.911257  \\
  \hline
\end{tabular}
\end{table*}

\subsection{Parameter Analysis}

\begin{figure}
\centering
\subfigure [Effect on Precision@5]{ \includegraphics[width=4cm]{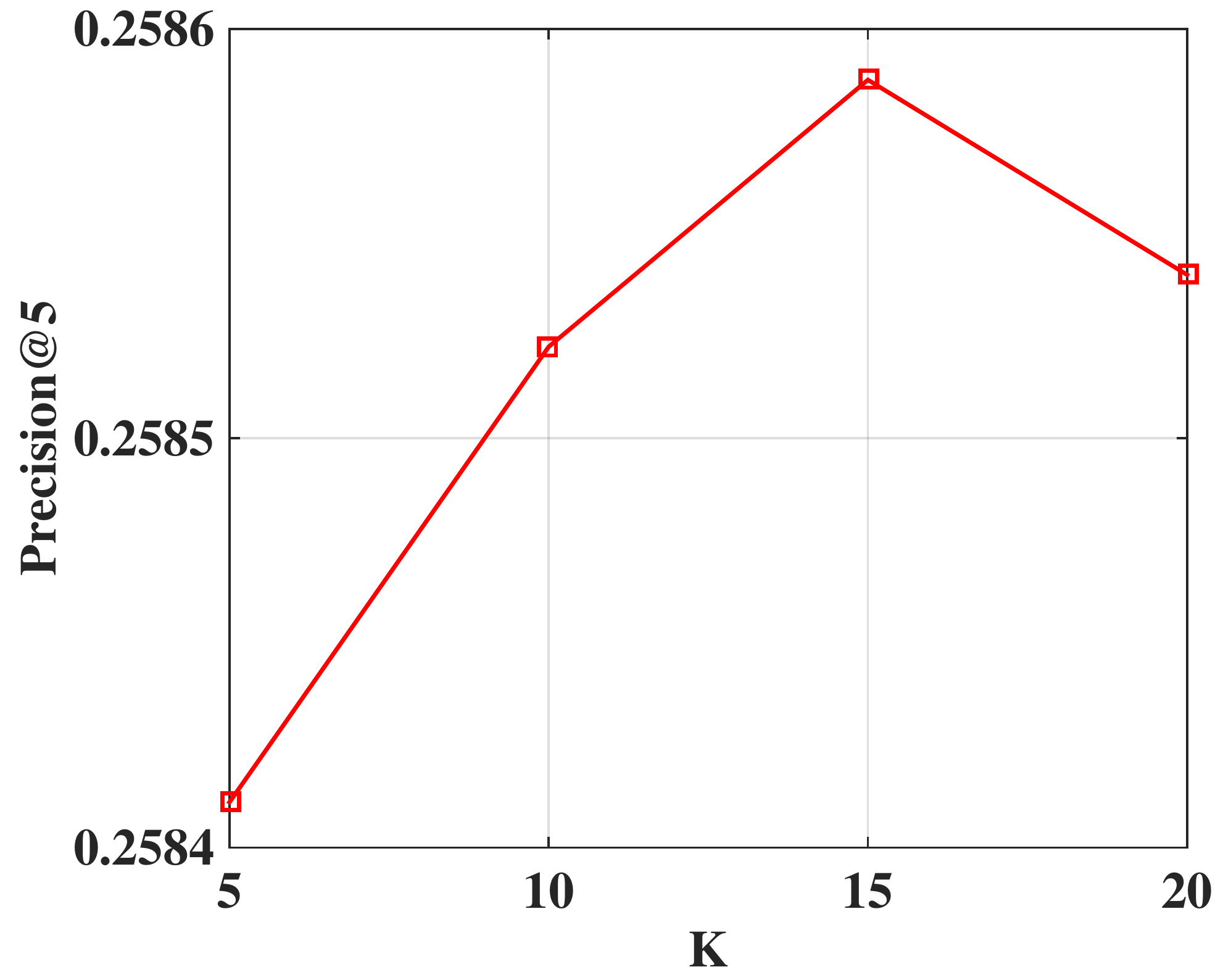}}~~~
\subfigure[Effect on DCG@5] { \includegraphics[width=4cm]{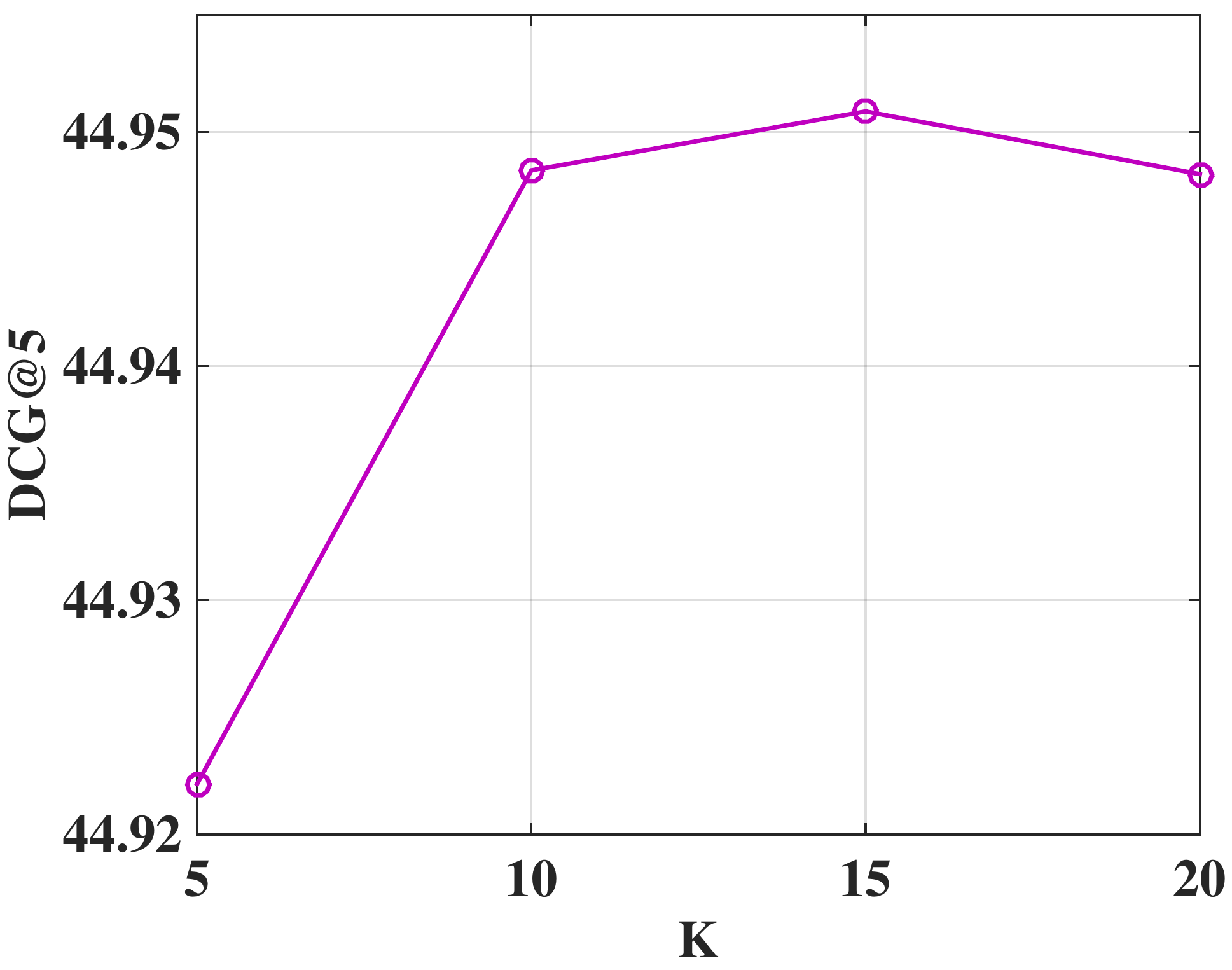}}
\caption{Effect of $K$ on DCH. Dataset used: \emph{Netflix90}.}
\label{effectk}
\end{figure}

Our model mainly has five hyperparameters, i.e., hash code length ($K$), hash code balanced degree ($\lambda$), worker number ($W$), minibatch size ($B$), and synchronization period ($P$).
We first fix $W=50$, $B=1000$, and $P=2$, and study the effects of $K$ and $\lambda$.

\textbf{Effect of $K$.} 
In the above sub-section, we have compared each model's performance with different $K$, we now study its effect on our model performance in details.
Figure \ref{effectk} shows the effect of $K$ on \emph{Netflix90} where $\lambda=0.01$.
We observe that our model performance first increases and then decreases after a certain value of $K$.
This is because the bigger $K$ is, the more information user and item hash codes contain.
Thus, our model performance increases with $K$ at first.
However, our model tends to be over-fitting when $K$ is too bigger, i.e., $K=15$. 
The best value of $K$ can be determined by cross-validation in practice.
In general, we find DCH achieves the best performance when $K$ is around 15 on \emph{Netflix}.

\begin{figure}
\centering
\subfigure [Effect on Precision@5]{ \includegraphics[width=4cm]{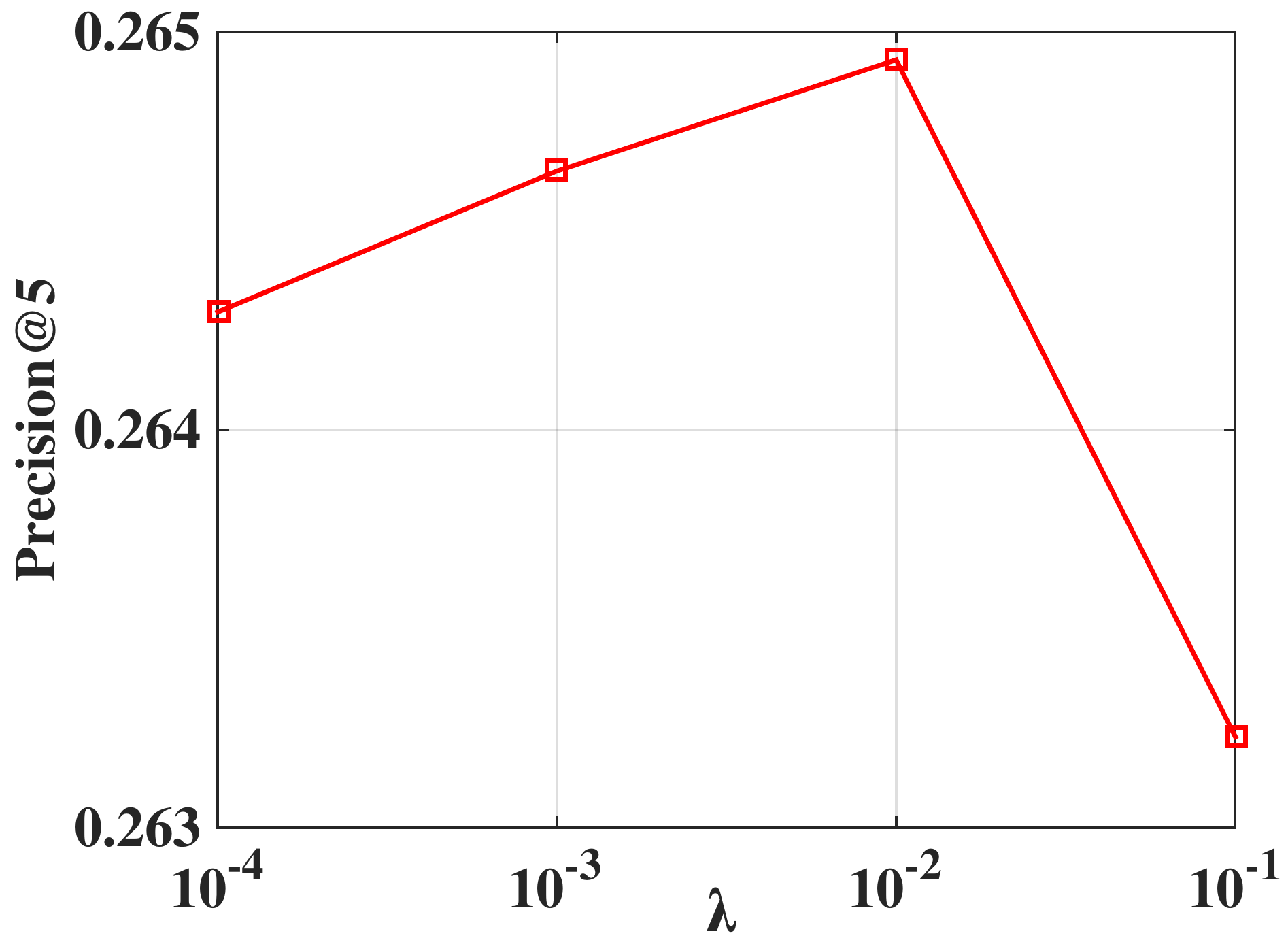}}~~~
\subfigure[Effect on DCG@5] { \includegraphics[width=4cm]{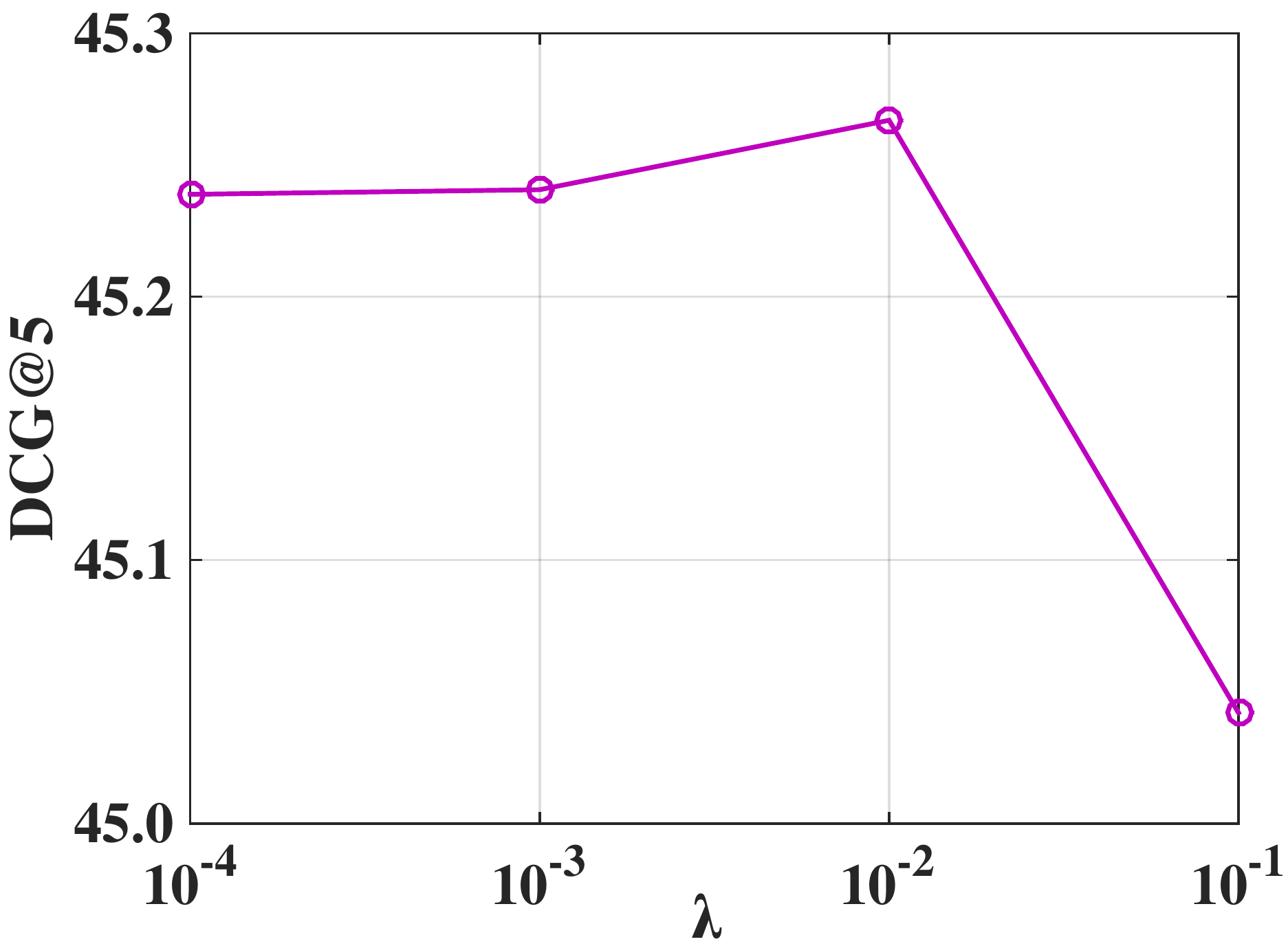}}
\caption{Effect of $\lambda$ on DCH. Dataset used: \emph{Netflix80}.}
\label{lambda}
\end{figure}

\textbf{Effect of $\lambda$.} 
Parameter $\lambda$ controls the balanced degree of user and item hash codes.
The bigger $\lambda$ is, the more balanced user and item hash codes are.
Figure \ref{lambda} shows the effect of $\lambda$ on \emph{Netflix80} where $K=10$.
As we can see, our model performance first increases and then decreases after a certain value of $\lambda$.
This indicates that the accuracy of DCH can be further improved with a good value of $\lambda$, since a better user and item hash codes are learnt.
The best value of $\lambda$ can also be determined by cross validation when using DCH.

We then fix $K=10$ and $\lambda=0.01$ and study the effects of $W$, $B$, and $P$.

\begin{figure}[!tp]
\centering
\subfigure [Effect of $W$]{ \includegraphics[width=3.9cm,height=3.15cm]{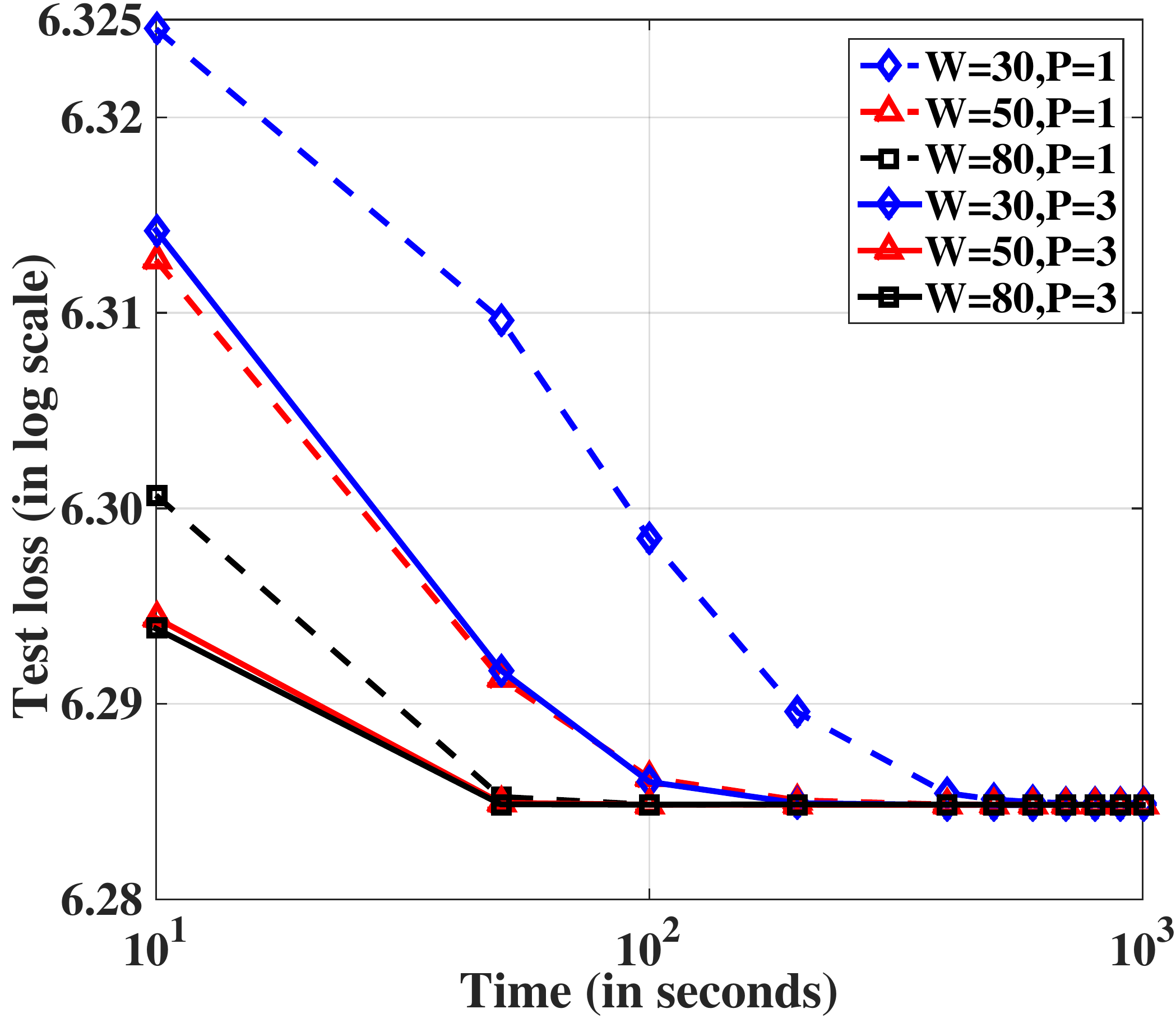}}
\subfigure [Effect of $B$]{ \includegraphics[width=3.9cm,height=3.15cm]{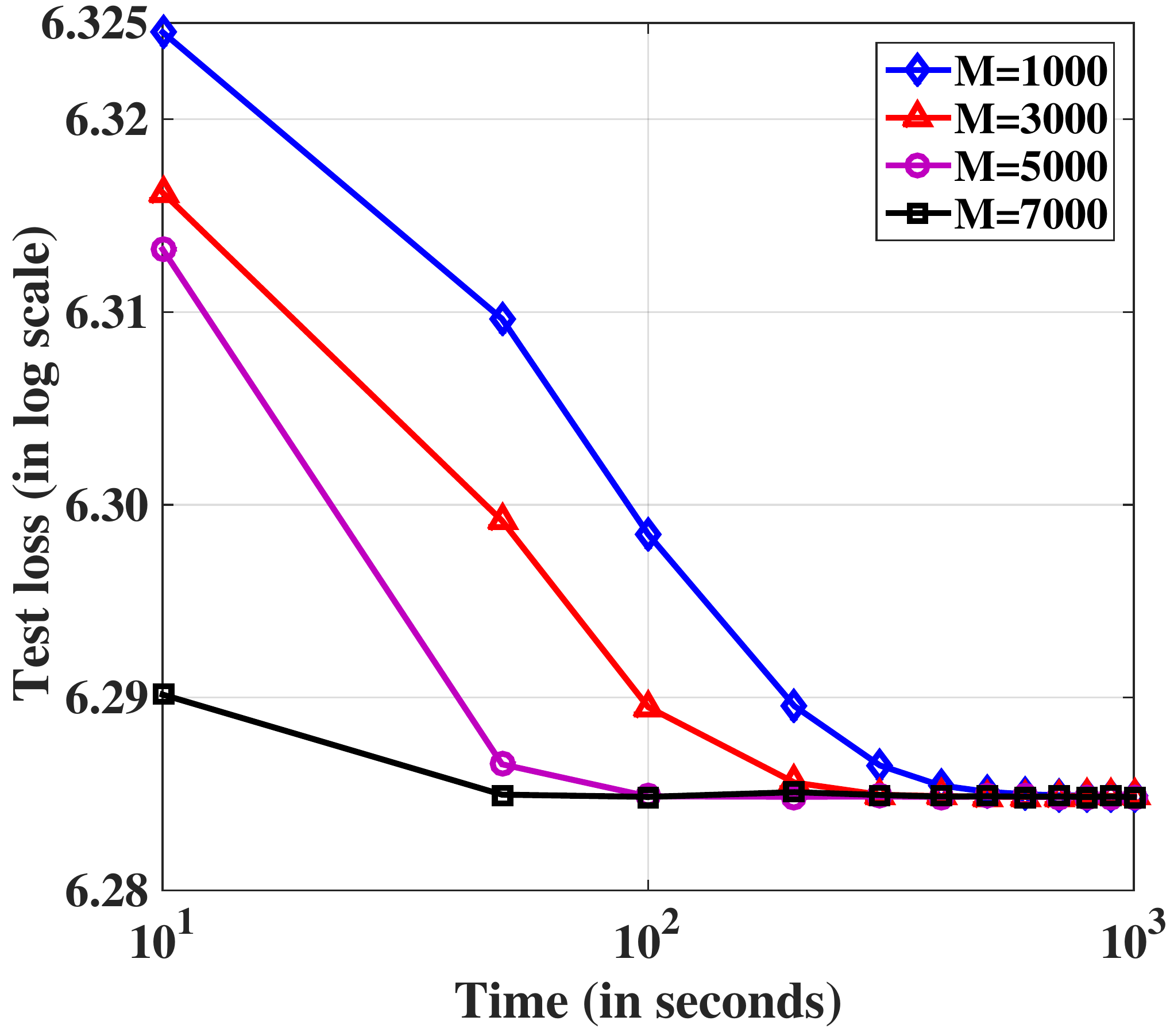}}
\caption{Effect of \#worker ($W$) and minibatch size ($B$) on model convergence speed. Dataset used: \emph{Netflix}.}
\label{workerloss}
\end{figure}

\textbf{Effect of $W$ and $B$.} 
Parameter $W$ controls the worker number, i.e., how many workers perform asynchronous SGD independently. 
Parameter $B$ controls the minibatch size, i.e., each worker loads and processes $B$ training data during each SGD operation. 
Both of them control the total size of training data workers can process in a certain epoch.
Figure \ref{workerloss} (a) shows the effect of $W$ on model convergence speed, where we fix $B=1000$, and set $P=1$ (dash lines) and $P=3$ (solid lines). 
Figure \ref{workerloss} (b) shows the effect of $B$ on model converge speed, where we fix $W=30$ and $P=1$. 
From them, we find that more workers and bigger minibatch size can speedup model convergence. 
The results are reasonable, because our model can process more training data by using more workers and increasing minibatch size, and thus converges faster. 

\begin{figure}[!tp]
\centering
\subfigure [Effect of $P$ on convergence speed]{ \includegraphics[width=3.9cm,height=3.05cm]{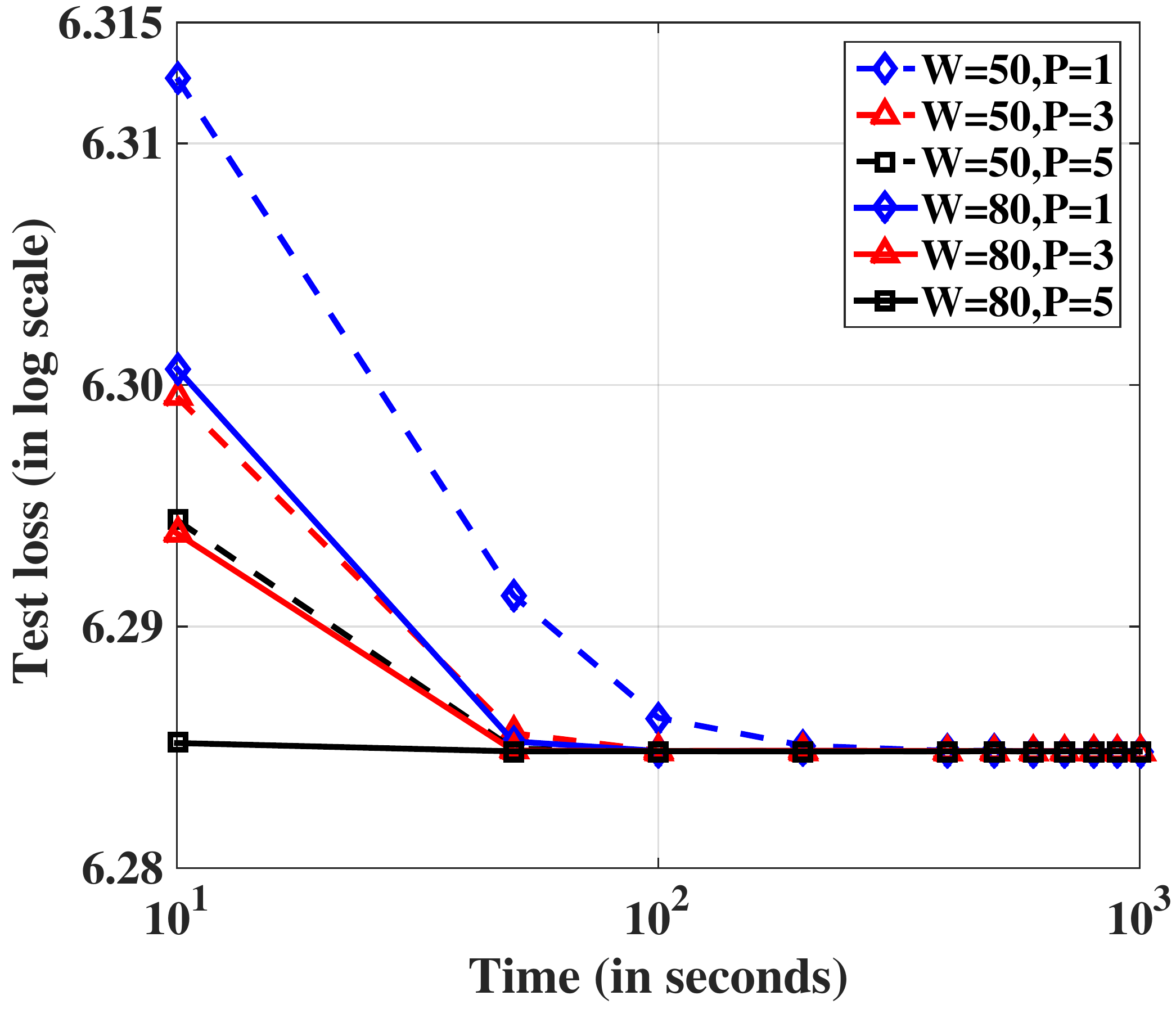}}
\subfigure [Effect of $P$ on model stability]{ \includegraphics[width=3.9cm,height=3.05cm]{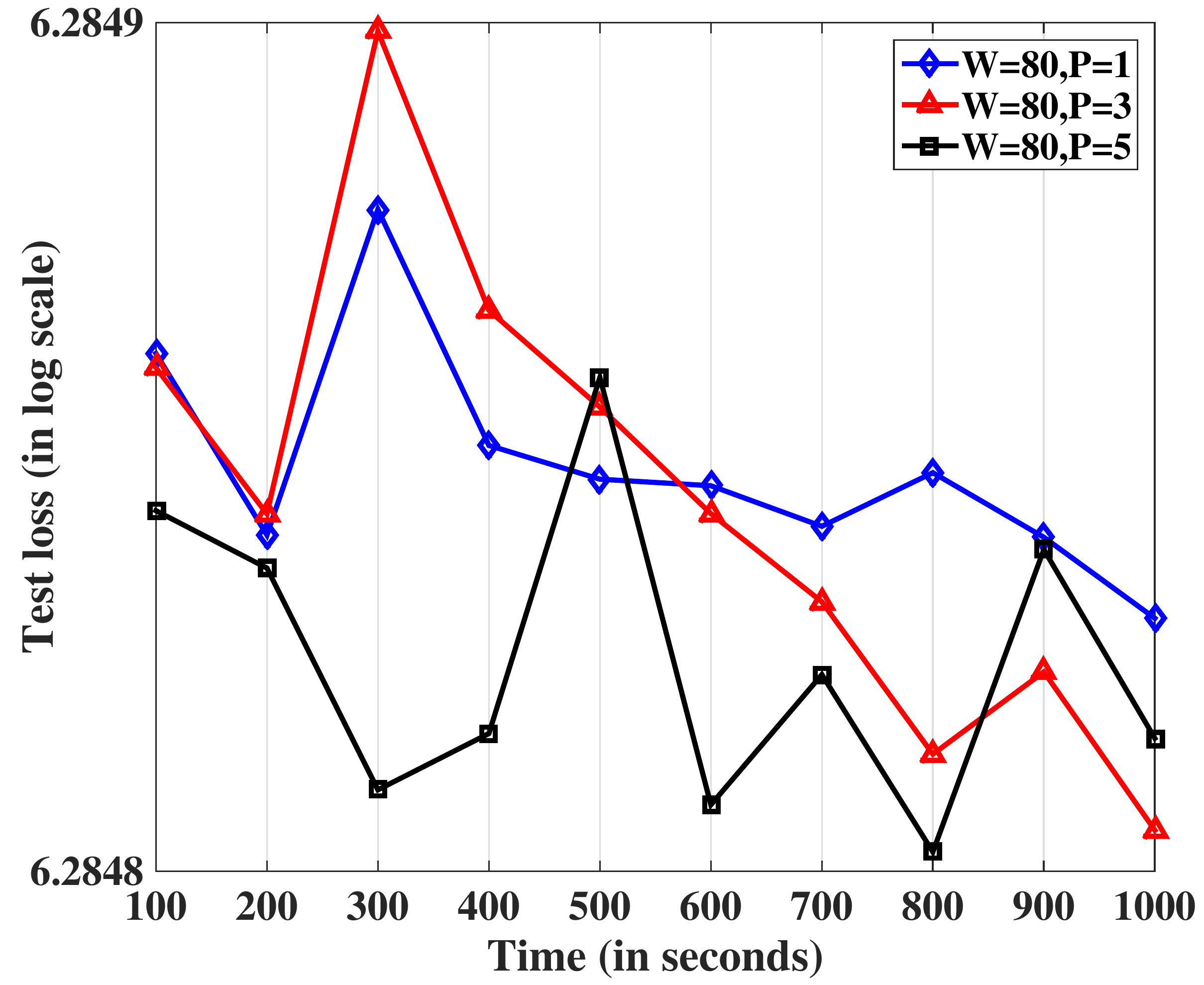}}
\caption{Effect of synchronization period ($P$) on model convergence. Dataset used: \emph{Netflix}.}
\label{nsyncloss}
\end{figure}

\textbf{Effect of $P$.} 
Parameter $P$ controls the period of synchronization, i.e., servers will wait for all the workers to synchronize once after each worker has done $P$ SGD operations. 
Figure \ref{nsyncloss} (a) shows the effect of $P$ on model convergence speed, where we fix $B=1000$, and use $W=50$ (dash lines) and $W=80$ (solid lines) respectively. 
From it, we find that our model converges faster with a bigger $P$. 
The speedup ratio is 10 when increasing $P$ from 1 to 5 by fixing other parameters.
Workers process SGD procedures independently without waiting for each other during the asynchronization periods, which makes the model converge much faster, especially when $P$ is big. 
However, we also calculate model variance and find that model becomes relative unstable when $P$ is too big. 
We show this finding in Figure \ref{nsyncloss} (b), where we fix $B=1000$ and $W=80$. 
As we explained in Section \ref{boundedstaleness}, each worker will probably not send the most recent model gradient to servers when using asynchronous SGD with bounded staleness, which may cause the model deviate too much. 
An appropriate value of $P$ (staleness bound) can not only make model converges faster, but also more stable. 
This experiment proves the practicalness of asynchronous SGD with bounded staleness. 

\subsection{Recommendation Efficiency}\label{efficiency}
First, we compare the offline model training efficiency of the classic latent factor model (i.e., MF) and DCH, where we set $B=50,000$ and $P=1$. 
Note that we do not compare DCH with DFM or DFMH here, because DFM, DFMH, and DCH are all distributed algorithms implemented on parameter server and share almost the same training speed.  
The results are shown in Table \ref{offlinetraining}: DCH significantly decreases the offline training time of MF. 
Besides, the more workers DCH uses, the bigger speedup of DCH against MF, since DCH can process more data with more workers and thus converges faster. 

\begin{table}\small
\centering
\caption{Comparison of offline training time (in seconds)}
\label{offlinetraining}
\begin{tabular}{|c|c|c|c|}
  \hline
  Model & MF & DCH (W=5) & DCH (W=10)  \\
  \hline
  \hline
  Training time & 8,461 & 2,142 & 1,351 \\
  \hline
  DCH speedup & - & 3.95 & 6.26 \\
  \hline
\end{tabular}
\end{table}

Next, we compare the online recommendation efficiency of the (distributed) latent factor models (i.e., MF and DFM) and DCH. 
We perform the following experiments on the \emph{Netflix} dataset, that is, we make recommendation from a pool of 17,770 candidate items. 
For MF and DFM which share the same online speed, we first compute user-item ratings using the learnt real-valued latent factors of user and item, and then sort the ratings, and finally return the top-10 items with highest ratings for each user. 
We name this approach ``real-valued rating rank'' and use ``real-valued'' for abbreviation. 
The time complexity of computing all the user-item ratings is $O(MNK)$, and the time complexity of sorting the top-k items is $O(MN)$.
For DCH, we choose Hamming distance search method as described in Section \ref{online} and return items within Hamming distance `1' to the binary codes of each user. 
We name it ``Hash'' for abbreviation, whose time complexity is $O(MN)$.
Note that this experiment is done on a Macbook Air with Intel 64 bit CPU and 8GB memory, and no parallel computing is involved.

\begin{figure}
\centering
\subfigure [Item size: 10000]{ \includegraphics[width=4cm]{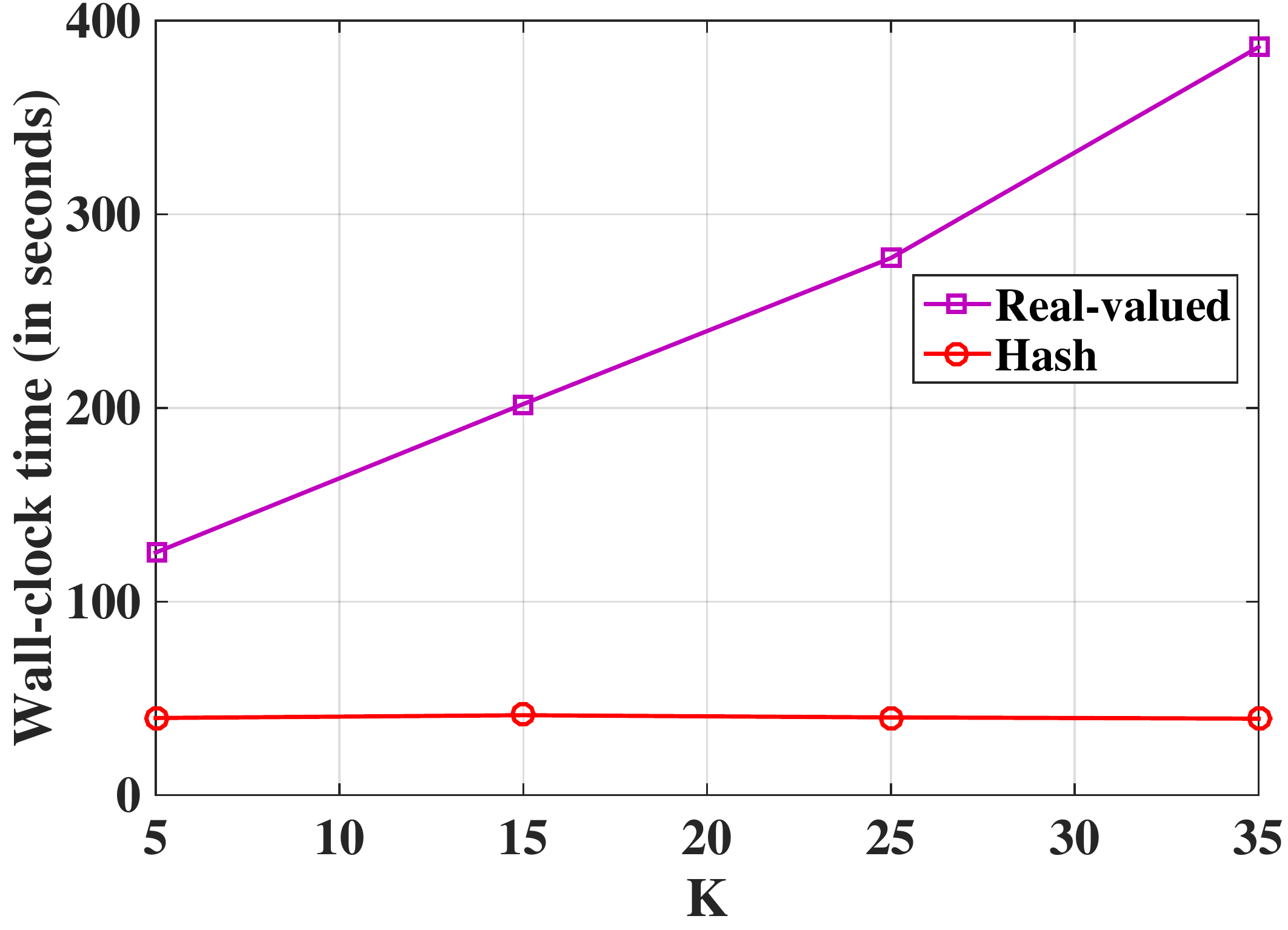}}
\subfigure [Item size: 17770] { \includegraphics[width=4cm]{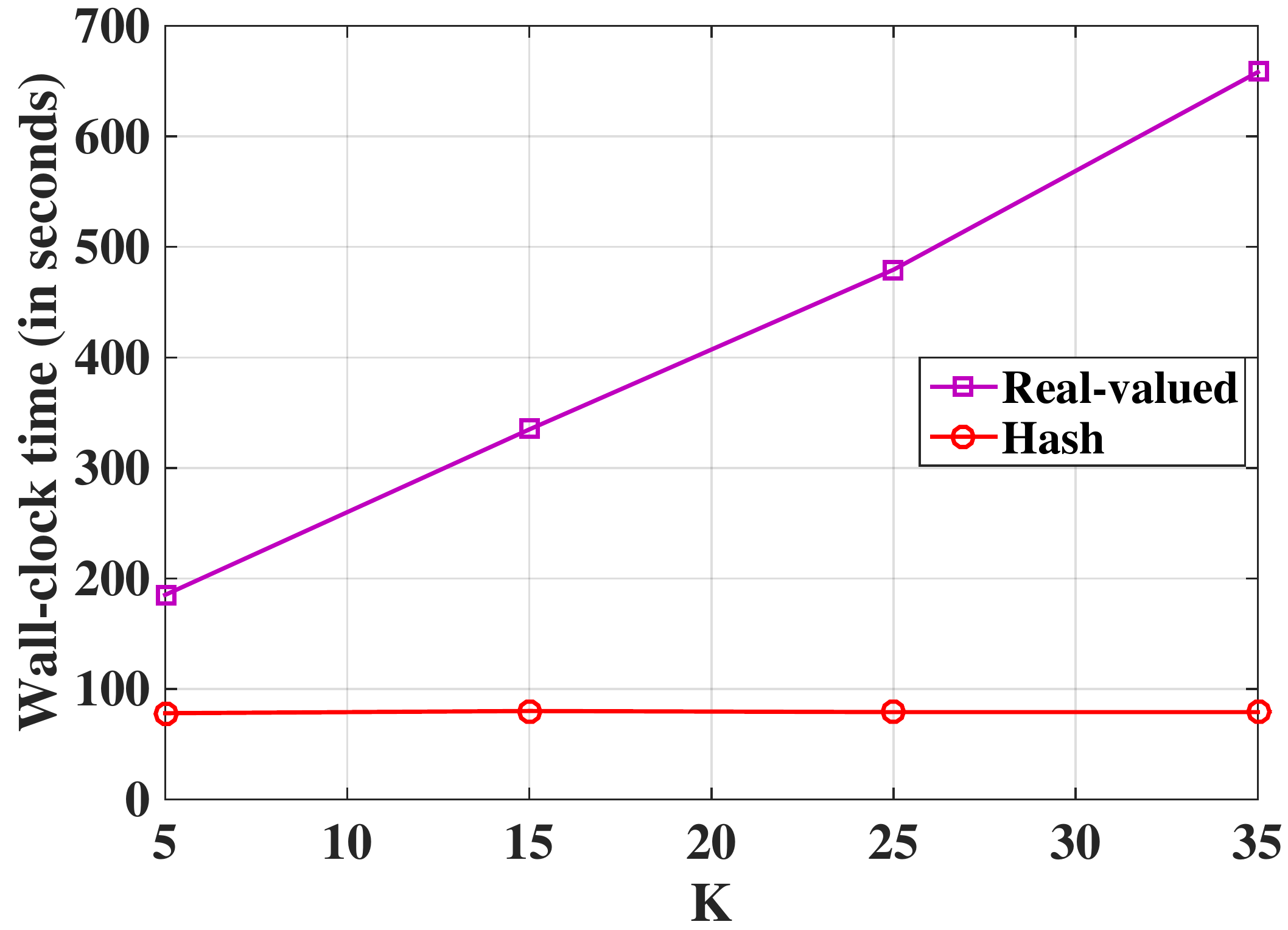}}
\caption{Time cost with different hash code length. }
\label{timek}
\end{figure}

\textbf{Efficiency in terms of hash code length.} 
Figure \ref{timek} shows the time cost of real-valued rating rank and Hamming distance search, where hash code length $K$ varies in $\{5, 15, 25, 35\}$ and item size is 17770. 
As we analyzed in Section \ref{online}, the time cost of real-valued rating rank scales linearly with the dimension of latent factors.
In contrast, the time cost of Hamming distance search is constant w.r.t the hash code length. 
For example, Hamming distance search only uses 80 seconds to make recommendation for 480,189 users. 
In other words, it takes only 0.17 milliseconds to make recommendation through 17,770 items for a single user, which totally meets the realtime requirement of online recommendation. 
These results demonstrate the effectiveness of hashing techniques. 

\begin{figure}
\centering
\subfigure [K=15]{ \includegraphics[width=4cm]{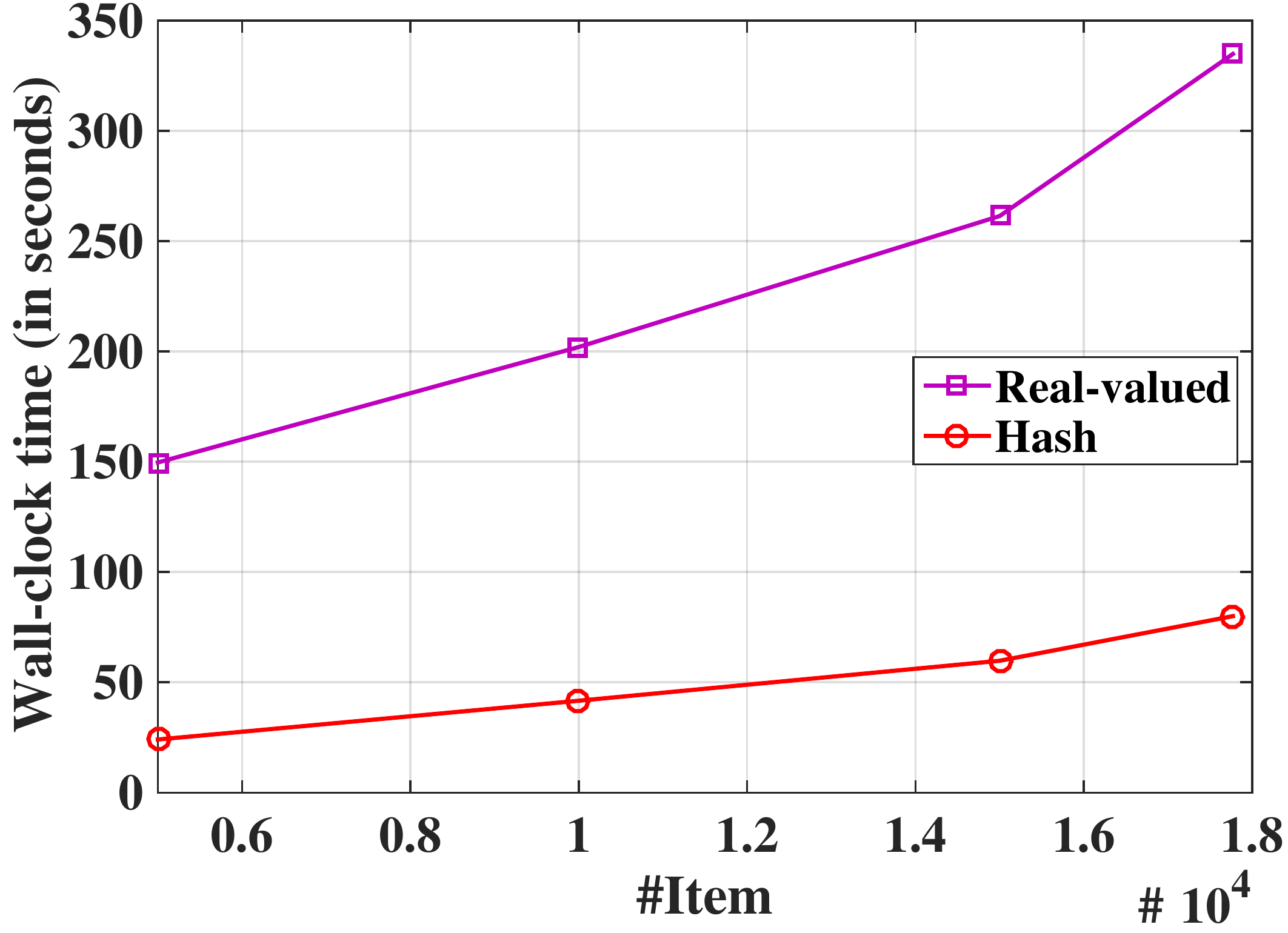}}
\subfigure [K=35] { \includegraphics[width=4cm]{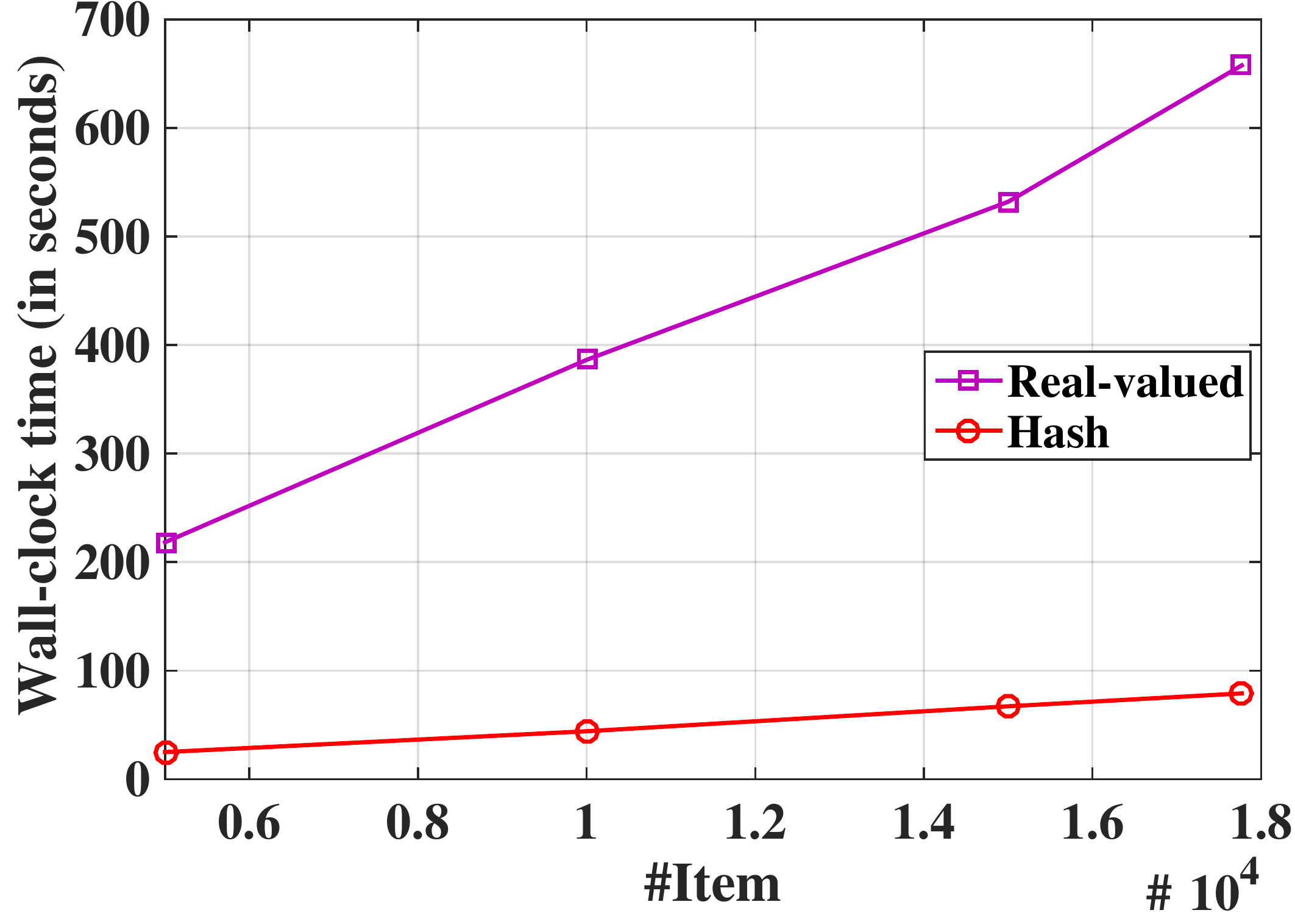}}
\caption{Time cost with different item number. }
\label{timeitem}
\end{figure}

\textbf{Efficiency in terms of item size.} 
Figure \ref{timeitem} shows the time cost of real-valued rating rank and Hamming distance search, where item size varies in $\{5000, 15000, 20000, 17770\}$ and $K=15$. 
From them, we can see that although both real-valued rating rank and Hamming distance search scales linearly with item size, their slopes are different. 
The time cost of real-valued rating rank increases much faster than Hamming distance search, especially when $K$ is large. 
This is because the time complexity of real-valued rating rank is $O(MNK)$, while the time complexity of Hamming distance search is $O(MN)$.
Moreover, one can also use multi-index hashing or other hashing techniques whose time complexity are constant w.r.t item size when the candidate item size is very large.
 
In summary, comparing with the traditional MF approaches, DCH is able to significantly improves efficiencies, including both offline model training efficiency and online recommendation efficiency, without losing much accuracy. 
With such a good property of DCH, it is suitable to be applied into real large-scale scenarios.

\subsection{Applications}
Our proposed DCH model has been successfully deployed into several real products at Ant Financial, e.g., the prize recommendation of scratch cards during the Chinese New Year 2017 and 
merchant recommendation in the ``Guess You Like'' scenario of Alipay. 
On one hand, the learnt hash codes provide additional useful features for both users and items, and thus can also be integrated into other models, e.g., logistic regression and deep neural network, to further improve their performances.
On the other hand, DCH is able to effectively retrieve user's latent interesting items, and thus can be used to match candidate items before ranking or re-ranking. 
The deployment of DCH can not only improve the click-through rate (CTR) of online recommendations by taking user and item hash codes an additional features, but also significantly decrease the matching time by using the hashing technique.

\section{Conclusion and Future Work}
In this paper, we aim to solve the efficiency problem of the existing collaborative models, including offline model training efficiency and online recommendation efficiency.
To do this, we proposed a Distributed Collaborative Hashing (DCH) model that aims to learn user and item hash codes from the existing user-item action history.
To improve offline model training efficiency, we designed DCH following the state-of-the-art parameter server paradigm. 
Specifically, we optimized DCH by distributedly computing subgradients on minibatches on workers asynchronously and updating model parameters on servers with bounded staleness. 
To improve online recommendation efficiency, we adopted hashing technique that has linear or even constant time complexity.
Finally, we conducted experiments on two large-scale datasets to prove our model performance and study parameter effects.
The results demonstrated that, comparing with the classic and state-of-the-art distributed collaborative models, 
(1) DCH has comparable performance in terms of recommendation accuracy, and
(2) DCH has fast convergence speed in offline model training  and realtime efficiency in online recommendation.

Our future work will focus on two directions.
One is distributed discrete hash code learning. 
Our current offline model training approach is two-stage, i.e., first solve the relaxed optimization problem and then rounding off.
Inspired by \cite{zhang2016discrete,lian2017discrete}, we plan to directly optimize user and item binary hash codes distributedly to further improve model performance. 
The other is to apply our model into graph representation technique for recommendation. As discussed in Section 3.6, we plan to hash code embeddings for nodes in graphs. 

\begin{acks}
We would like to acknowledge contributions from our colleagues from Alibaba and Ant Financial, including: Xu Chen, Yi Ding, Qing Cui, Jin Yu, and Jian Huang. We would also like to extend our sincere thanks to the entire Large-Scale-Learning and Distributed-Learning-and-Systems team members. Finally, we thank KDD anonymous reviewers for their helpful suggestions.
\end{acks}

\bibliographystyle{ACM-Reference-Format}
\balance
\bibliography{hashmf-reference} 


\begin{thebibliography}{39}


\ifx \showCODEN    \undefined \def \showCODEN     #1{\unskip}     \fi
\ifx \showDOI      \undefined \def \showDOI       #1{#1}\fi
\ifx \showISBNx    \undefined \def \showISBNx     #1{\unskip}     \fi
\ifx \showISBNxiii \undefined \def \showISBNxiii  #1{\unskip}     \fi
\ifx \showISSN     \undefined \def \showISSN      #1{\unskip}     \fi
\ifx \showLCCN     \undefined \def \showLCCN      #1{\unskip}     \fi
\ifx \shownote     \undefined \def \shownote      #1{#1}          \fi
\ifx \showarticletitle \undefined \def \showarticletitle #1{#1}   \fi
\ifx \showURL      \undefined \def \showURL       {\relax}        \fi
\providecommand\bibfield[2]{#2}
\providecommand\bibinfo[2]{#2}
\providecommand\natexlab[1]{#1}
\providecommand\showeprint[2][]{arXiv:#2}

\bibitem[\protect\citeauthoryear{Agarwal and Chen}{Agarwal and Chen}{2009}]%
        {agarwal2009regression}
\bibfield{author}{\bibinfo{person}{Deepak Agarwal} {and}
  \bibinfo{person}{Bee-Chung Chen}.} \bibinfo{year}{2009}\natexlab{}.
\newblock \showarticletitle{Regression-based latent factor models}. In
  \bibinfo{booktitle}{\emph{Proceedings of the 14th ACM SIGKDD International
  Conference on Knowledge Discovery and Data Mining}}. \bibinfo{pages}{19--28}.
\newblock


\bibitem[\protect\citeauthoryear{Bennett and Lanning}{Bennett and
  Lanning}{2007}]%
        {bennett2007netflix}
\bibfield{author}{\bibinfo{person}{James Bennett} {and} \bibinfo{person}{Stan
  Lanning}.} \bibinfo{year}{2007}\natexlab{}.
\newblock \showarticletitle{The netflix prize}. In
  \bibinfo{booktitle}{\emph{Proceedings of KDD Cup and Workshop 2007 Aug 12}},
  Vol.~\bibinfo{volume}{2007}. \bibinfo{pages}{35}.
\newblock


\bibitem[\protect\citeauthoryear{Cao, Lu, and Xu}{Cao et~al\mbox{.}}{2015}]%
        {cao2015grarep}
\bibfield{author}{\bibinfo{person}{Shaosheng Cao}, \bibinfo{person}{Wei Lu},
  {and} \bibinfo{person}{Qiongkai Xu}.} \bibinfo{year}{2015}\natexlab{}.
\newblock \showarticletitle{Grarep: Learning graph representations with global
  structural information}. In \bibinfo{booktitle}{\emph{Proceedings of the 24th
  ACM International on Conference on Information and Knowledge Management}}.
  ACM, \bibinfo{pages}{891--900}.
\newblock


\bibitem[\protect\citeauthoryear{Chen, Zheng, Wang, Hong, and Lin}{Chen
  et~al\mbox{.}}{2014}]%
        {chen2014context}
\bibfield{author}{\bibinfo{person}{Chaochao Chen}, \bibinfo{person}{Xiaolin
  Zheng}, \bibinfo{person}{Yan Wang}, \bibinfo{person}{Fuxing Hong}, {and}
  \bibinfo{person}{Zhen Lin}.} \bibinfo{year}{2014}\natexlab{}.
\newblock \showarticletitle{Context-aware Collaborative Topic Regression with
  Social Matrix Factorization for Recommender Systems}.
\newblock In \bibinfo{booktitle}{\emph{AAAI}}. \bibinfo{pages}{9--15}.
\newblock


\bibitem[\protect\citeauthoryear{Chen, Zheng, Zhu, and Xiao}{Chen
  et~al\mbox{.}}{2016}]%
        {chen2016recommender}
\bibfield{author}{\bibinfo{person}{Chaochao Chen}, \bibinfo{person}{Xiaolin
  Zheng}, \bibinfo{person}{Mengying Zhu}, {and} \bibinfo{person}{Litao Xiao}.}
  \bibinfo{year}{2016}\natexlab{}.
\newblock \showarticletitle{Recommender System with Composite Social Trust
  Networks}.
\newblock \bibinfo{journal}{\emph{International Journal of Web Services
  Research (IJWSR)}} \bibinfo{volume}{13}, \bibinfo{number}{2}
  (\bibinfo{year}{2016}), \bibinfo{pages}{56--73}.
\newblock


\bibitem[\protect\citeauthoryear{Das, Datar, Garg, and Rajaram}{Das
  et~al\mbox{.}}{2007}]%
        {das2007google}
\bibfield{author}{\bibinfo{person}{Abhinandan~S Das}, \bibinfo{person}{Mayur
  Datar}, \bibinfo{person}{Ashutosh Garg}, {and} \bibinfo{person}{Shyam
  Rajaram}.} \bibinfo{year}{2007}\natexlab{}.
\newblock \showarticletitle{Google news personalization: scalable online
  collaborative filtering}. In \bibinfo{booktitle}{\emph{Proceedings of the
  16th International Conference on World Wide Web}}. ACM,
  \bibinfo{pages}{271--280}.
\newblock


\bibitem[\protect\citeauthoryear{Eksombatchai, Jindal, Liu, Liu, Sharma,
  Sugnet, Ulrich, and Leskovec}{Eksombatchai et~al\mbox{.}}{2018}]%
        {eksombatchai2017pixie}
\bibfield{author}{\bibinfo{person}{Chantat Eksombatchai},
  \bibinfo{person}{Pranav Jindal}, \bibinfo{person}{Jerry~Zitao Liu},
  \bibinfo{person}{Yuchen Liu}, \bibinfo{person}{Rahul Sharma},
  \bibinfo{person}{Charles Sugnet}, \bibinfo{person}{Mark Ulrich}, {and}
  \bibinfo{person}{Jure Leskovec}.} \bibinfo{year}{2018}\natexlab{}.
\newblock \showarticletitle{Pixie: A System for Recommending 3+ Billion Items
  to 200+ Million Users in Real-Time}. In \bibinfo{booktitle}{\emph{Proceedings
  of the 27th International Conference on World Wide Web}}.
\newblock


\bibitem[\protect\citeauthoryear{Gemulla, Nijkamp, Haas, and Sismanis}{Gemulla
  et~al\mbox{.}}{2011}]%
        {gemulla2011large}
\bibfield{author}{\bibinfo{person}{Rainer Gemulla}, \bibinfo{person}{Erik
  Nijkamp}, \bibinfo{person}{Peter~J Haas}, {and} \bibinfo{person}{Yannis
  Sismanis}.} \bibinfo{year}{2011}\natexlab{}.
\newblock \showarticletitle{Large-scale matrix factorization with distributed
  stochastic gradient descent}. In \bibinfo{booktitle}{\emph{Proceedings of the
  17th ACM SIGKDD International Conference on Knowledge Discovery and Data
  Mining}}. ACM, \bibinfo{pages}{69--77}.
\newblock


\bibitem[\protect\citeauthoryear{Grover and Leskovec}{Grover and
  Leskovec}{2016}]%
        {grover2016node2vec}
\bibfield{author}{\bibinfo{person}{Aditya Grover} {and} \bibinfo{person}{Jure
  Leskovec}.} \bibinfo{year}{2016}\natexlab{}.
\newblock \showarticletitle{node2vec: Scalable feature learning for networks}.
  In \bibinfo{booktitle}{\emph{Proceedings of the 22nd ACM SIGKDD international
  conference on Knowledge discovery and data mining}}. ACM,
  \bibinfo{pages}{855--864}.
\newblock


\bibitem[\protect\citeauthoryear{Karatzoglou, Weimer, and Smola}{Karatzoglou
  et~al\mbox{.}}{2010}]%
        {karatzoglou2010collaborative}
\bibfield{author}{\bibinfo{person}{Alexandros Karatzoglou},
  \bibinfo{person}{Markus Weimer}, {and} \bibinfo{person}{Alex~J Smola}.}
  \bibinfo{year}{2010}\natexlab{}.
\newblock \showarticletitle{Collaborative filtering on a budget}. In
  \bibinfo{booktitle}{\emph{Proceedings of the Thirteenth International
  Conference on Artificial Intelligence and Statistics}}.
  \bibinfo{pages}{389--396}.
\newblock


\bibitem[\protect\citeauthoryear{Koren}{Koren}{2008}]%
        {koren2008factorization}
\bibfield{author}{\bibinfo{person}{Yehuda Koren}.}
  \bibinfo{year}{2008}\natexlab{}.
\newblock \showarticletitle{Factorization meets the neighborhood: a
  multifaceted collaborative filtering model}. In
  \bibinfo{booktitle}{\emph{Proceedings of the 15th ACM SIGKDD International
  Conference on Knowledge Discovery and Data Mining}}.
  \bibinfo{pages}{426--434}.
\newblock


\bibitem[\protect\citeauthoryear{Koren, Bell, Volinsky, et~al\mbox{.}}{Koren
  et~al\mbox{.}}{2009}]%
        {koren2009matrix}
\bibfield{author}{\bibinfo{person}{Yehuda Koren}, \bibinfo{person}{Robert
  Bell}, \bibinfo{person}{Chris Volinsky}, {et~al\mbox{.}}}
  \bibinfo{year}{2009}\natexlab{}.
\newblock \showarticletitle{Matrix factorization techniques for recommender
  systems}.
\newblock \bibinfo{journal}{\emph{Computer}} \bibinfo{volume}{42},
  \bibinfo{number}{8} (\bibinfo{year}{2009}), \bibinfo{pages}{30--37}.
\newblock


\bibitem[\protect\citeauthoryear{Kulis and Grauman}{Kulis and Grauman}{2009}]%
        {kulis2009kernelized}
\bibfield{author}{\bibinfo{person}{Brian Kulis} {and} \bibinfo{person}{Kristen
  Grauman}.} \bibinfo{year}{2009}\natexlab{}.
\newblock \showarticletitle{Kernelized locality-sensitive hashing for scalable
  image search}. In \bibinfo{booktitle}{\emph{Proceedings of the 12th
  International Conference on Computer Vision}}. IEEE,
  \bibinfo{pages}{2130--2137}.
\newblock


\bibitem[\protect\citeauthoryear{Li, Andersen, Park, Smola, Ahmed, Josifovski,
  Long, Shekita, and Su}{Li et~al\mbox{.}}{2014}]%
        {li2014scaling}
\bibfield{author}{\bibinfo{person}{Mu Li}, \bibinfo{person}{David~G Andersen},
  \bibinfo{person}{Jun~Woo Park}, \bibinfo{person}{Alexander~J Smola},
  \bibinfo{person}{Amr Ahmed}, \bibinfo{person}{Vanja Josifovski},
  \bibinfo{person}{James Long}, \bibinfo{person}{Eugene~J Shekita}, {and}
  \bibinfo{person}{Bor-Yiing Su}.} \bibinfo{year}{2014}\natexlab{}.
\newblock \showarticletitle{Scaling distributed machine learning with the
  parameter server}. In \bibinfo{booktitle}{\emph{OSDI}}.
  \bibinfo{pages}{583--598}.
\newblock


\bibitem[\protect\citeauthoryear{Li, Liu, Smola, and Wang}{Li
  et~al\mbox{.}}{2016}]%
        {li2016difacto}
\bibfield{author}{\bibinfo{person}{Mu Li}, \bibinfo{person}{Ziqi Liu},
  \bibinfo{person}{Alexander~J Smola}, {and} \bibinfo{person}{Yu-Xiang Wang}.}
  \bibinfo{year}{2016}\natexlab{}.
\newblock \showarticletitle{DiFacto: Distributed Factorization Machines}. In
  \bibinfo{booktitle}{\emph{Proceedings of the Ninth ACM International
  Conference on Web Search and Data Mining}}. ACM, \bibinfo{pages}{377--386}.
\newblock


\bibitem[\protect\citeauthoryear{Lian, Liu, Ge, Zheng, Xie, and Cao}{Lian
  et~al\mbox{.}}{2017}]%
        {lian2017discrete}
\bibfield{author}{\bibinfo{person}{Defu Lian}, \bibinfo{person}{Rui Liu},
  \bibinfo{person}{Yong Ge}, \bibinfo{person}{Kai Zheng}, \bibinfo{person}{Xing
  Xie}, {and} \bibinfo{person}{Longbing Cao}.} \bibinfo{year}{2017}\natexlab{}.
\newblock \showarticletitle{Discrete Content-aware Matrix Factorization}. In
  \bibinfo{booktitle}{\emph{Proceedings of the 23rd ACM SIGKDD International
  Conference on Knowledge Discovery and Data Mining}}. ACM,
  \bibinfo{pages}{325--334}.
\newblock


\bibitem[\protect\citeauthoryear{Liu, Wang, Ji, Jiang, and Chang}{Liu
  et~al\mbox{.}}{2012}]%
        {liu2012supervised}
\bibfield{author}{\bibinfo{person}{Wei Liu}, \bibinfo{person}{Jun Wang},
  \bibinfo{person}{Rongrong Ji}, \bibinfo{person}{Yu-Gang Jiang}, {and}
  \bibinfo{person}{Shih-Fu Chang}.} \bibinfo{year}{2012}\natexlab{}.
\newblock \showarticletitle{Supervised hashing with kernels}. In
  \bibinfo{booktitle}{\emph{Proceedings of the IEEE Conference on Computer
  Vision and Pattern Recognition}}. IEEE, \bibinfo{pages}{2074--2081}.
\newblock


\bibitem[\protect\citeauthoryear{Liu, He, Deng, and Lang}{Liu
  et~al\mbox{.}}{2014}]%
        {liu2014collaborative}
\bibfield{author}{\bibinfo{person}{Xianglong Liu}, \bibinfo{person}{Junfeng
  He}, \bibinfo{person}{Cheng Deng}, {and} \bibinfo{person}{Bo Lang}.}
  \bibinfo{year}{2014}\natexlab{}.
\newblock \showarticletitle{Collaborative hashing}. In
  \bibinfo{booktitle}{\emph{Proceedings of the IEEE Conference on Computer
  Vision and Pattern Recognition}}. \bibinfo{pages}{2139--2146}.
\newblock


\bibitem[\protect\citeauthoryear{McAuley and Leskovec}{McAuley and
  Leskovec}{2013}]%
        {mcauley2013hidden}
\bibfield{author}{\bibinfo{person}{Julian McAuley} {and} \bibinfo{person}{Jure
  Leskovec}.} \bibinfo{year}{2013}\natexlab{}.
\newblock \showarticletitle{Hidden factors and hidden topics: understanding
  rating dimensions with review text}. In \bibinfo{booktitle}{\emph{Proceedings
  of the 7th ACM conference on Recommender systems}}.
  \bibinfo{pages}{165--172}.
\newblock


\bibitem[\protect\citeauthoryear{Mnih and Salakhutdinov}{Mnih and
  Salakhutdinov}{2007}]%
        {mnih2007probabilistic}
\bibfield{author}{\bibinfo{person}{Andriy Mnih} {and} \bibinfo{person}{Ruslan
  Salakhutdinov}.} \bibinfo{year}{2007}\natexlab{}.
\newblock \showarticletitle{Probabilistic matrix factorization}. In
  \bibinfo{booktitle}{\emph{Advances in Neural Information Processing
  Systems}}. \bibinfo{pages}{1257--1264}.
\newblock


\bibitem[\protect\citeauthoryear{Norouzi, Punjani, and Fleet}{Norouzi
  et~al\mbox{.}}{2012}]%
        {norouzi2012fast}
\bibfield{author}{\bibinfo{person}{Mohammad Norouzi}, \bibinfo{person}{Ali
  Punjani}, {and} \bibinfo{person}{David~J Fleet}.}
  \bibinfo{year}{2012}\natexlab{}.
\newblock \showarticletitle{Fast search in hamming space with multi-index
  hashing}. In \bibinfo{booktitle}{\emph{Proceedings of the IEEE Conference on
  Computer Vision and Pattern Recognition (CVPR)}}. IEEE,
  \bibinfo{pages}{3108--3115}.
\newblock


\bibitem[\protect\citeauthoryear{Ou, Cui, Wang, Wang, Zhu, and Yang}{Ou
  et~al\mbox{.}}{2013}]%
        {ou2013comparing}
\bibfield{author}{\bibinfo{person}{Mingdong Ou}, \bibinfo{person}{Peng Cui},
  \bibinfo{person}{Fei Wang}, \bibinfo{person}{Jun Wang},
  \bibinfo{person}{Wenwu Zhu}, {and} \bibinfo{person}{Shiqiang Yang}.}
  \bibinfo{year}{2013}\natexlab{}.
\newblock \showarticletitle{Comparing apples to oranges: a scalable solution
  with heterogeneous hashing}. In \bibinfo{booktitle}{\emph{Proceedings of the
  19th ACM SIGKDD International Conference on Knowledge Discovery and Data
  Mining}}. ACM, \bibinfo{pages}{230--238}.
\newblock


\bibitem[\protect\citeauthoryear{Perozzi, Al-Rfou, and Skiena}{Perozzi
  et~al\mbox{.}}{2014}]%
        {perozzi2014deepwalk}
\bibfield{author}{\bibinfo{person}{Bryan Perozzi}, \bibinfo{person}{Rami
  Al-Rfou}, {and} \bibinfo{person}{Steven Skiena}.}
  \bibinfo{year}{2014}\natexlab{}.
\newblock \showarticletitle{Deepwalk: Online learning of social
  representations}. In \bibinfo{booktitle}{\emph{Proceedings of the 20th ACM
  SIGKDD international conference on Knowledge discovery and data mining}}.
  ACM, \bibinfo{pages}{701--710}.
\newblock


\bibitem[\protect\citeauthoryear{Ray and Koopman}{Ray and Koopman}{2006}]%
        {ray2006efficient}
\bibfield{author}{\bibinfo{person}{Justin Ray} {and} \bibinfo{person}{Philip
  Koopman}.} \bibinfo{year}{2006}\natexlab{}.
\newblock \showarticletitle{Efficient high hamming distance CRCs for embedded
  networks}. In \bibinfo{booktitle}{\emph{DSN}}. IEEE, \bibinfo{pages}{3--12}.
\newblock


\bibitem[\protect\citeauthoryear{Rendle}{Rendle}{2010}]%
        {rendle2010factorization}
\bibfield{author}{\bibinfo{person}{Steffen Rendle}.}
  \bibinfo{year}{2010}\natexlab{}.
\newblock \showarticletitle{Factorization machines}. In
  \bibinfo{booktitle}{\emph{Proceedings of the IEEE 10th International
  Conference on Data Mining (ICDM)}}. \bibinfo{pages}{995--1000}.
\newblock


\bibitem[\protect\citeauthoryear{Resnick, Iacovou, Suchak, Bergstrom, and
  Riedl}{Resnick et~al\mbox{.}}{1994}]%
        {resnick1994grouplens}
\bibfield{author}{\bibinfo{person}{Paul Resnick}, \bibinfo{person}{Neophytos
  Iacovou}, \bibinfo{person}{Mitesh Suchak}, \bibinfo{person}{Peter Bergstrom},
  {and} \bibinfo{person}{John Riedl}.} \bibinfo{year}{1994}\natexlab{}.
\newblock \showarticletitle{GroupLens: an open architecture for collaborative
  filtering of netnews}. In \bibinfo{booktitle}{\emph{Proceedings of the 1994
  ACM Conference on Computer Supported Cooperative Work}}.
  \bibinfo{pages}{175--186}.
\newblock


\bibitem[\protect\citeauthoryear{Salakhutdinov and Hinton}{Salakhutdinov and
  Hinton}{2009}]%
        {salakhutdinov2009semantic}
\bibfield{author}{\bibinfo{person}{Ruslan Salakhutdinov} {and}
  \bibinfo{person}{Geoffrey Hinton}.} \bibinfo{year}{2009}\natexlab{}.
\newblock \showarticletitle{Semantic hashing}.
\newblock \bibinfo{journal}{\emph{International Journal of Approximate
  Reasoning}} \bibinfo{volume}{50}, \bibinfo{number}{7} (\bibinfo{year}{2009}),
  \bibinfo{pages}{969--978}.
\newblock


\bibitem[\protect\citeauthoryear{Sarwar, Karypis, Konstan, and Riedl}{Sarwar
  et~al\mbox{.}}{2001}]%
        {sarwar2001item}
\bibfield{author}{\bibinfo{person}{Badrul Sarwar}, \bibinfo{person}{George
  Karypis}, \bibinfo{person}{Joseph Konstan}, {and} \bibinfo{person}{John
  Riedl}.} \bibinfo{year}{2001}\natexlab{}.
\newblock \showarticletitle{Item-based collaborative filtering recommendation
  algorithms}. In \bibinfo{booktitle}{\emph{WWW}}. \bibinfo{pages}{285--295}.
\newblock


\bibitem[\protect\citeauthoryear{Schelter, Satuluri, and Zadeh}{Schelter
  et~al\mbox{.}}{2014}]%
        {schelter2014factorbird}
\bibfield{author}{\bibinfo{person}{Sebastian Schelter}, \bibinfo{person}{Venu
  Satuluri}, {and} \bibinfo{person}{Reza Zadeh}.}
  \bibinfo{year}{2014}\natexlab{}.
\newblock \showarticletitle{Factorbird-a parameter server approach to
  distributed matrix factorization}.
\newblock \bibinfo{journal}{\emph{arXiv preprint arXiv:1411.0602}}
  (\bibinfo{year}{2014}).
\newblock


\bibitem[\protect\citeauthoryear{Shalev-Shwartz, Singer, Srebro, and
  Cotter}{Shalev-Shwartz et~al\mbox{.}}{2011}]%
        {shalev2011pegasos}
\bibfield{author}{\bibinfo{person}{Shai Shalev-Shwartz}, \bibinfo{person}{Yoram
  Singer}, \bibinfo{person}{Nathan Srebro}, {and} \bibinfo{person}{Andrew
  Cotter}.} \bibinfo{year}{2011}\natexlab{}.
\newblock \showarticletitle{Pegasos: Primal estimated sub-gradient solver for
  svm}.
\newblock \bibinfo{journal}{\emph{Mathematical Programming}}
  \bibinfo{volume}{127}, \bibinfo{number}{1} (\bibinfo{year}{2011}),
  \bibinfo{pages}{3--30}.
\newblock


\bibitem[\protect\citeauthoryear{Smirnov and Ponomarev}{Smirnov and
  Ponomarev}{2015}]%
        {smirnov2015locality}
\bibfield{author}{\bibinfo{person}{Alexander Smirnov} {and}
  \bibinfo{person}{Andrew Ponomarev}.} \bibinfo{year}{2015}\natexlab{}.
\newblock \showarticletitle{Locality-Sensitive Hashing for Distributed
  Privacy-Preserving Collaborative Filtering: An Approach and System
  Architecture}. In \bibinfo{booktitle}{\emph{International Conference on
  Enterprise Information Systems}}. Springer, \bibinfo{pages}{455--475}.
\newblock


\bibitem[\protect\citeauthoryear{Su and Khoshgoftaar}{Su and
  Khoshgoftaar}{2009}]%
        {su2009survey}
\bibfield{author}{\bibinfo{person}{Xiaoyuan Su} {and} \bibinfo{person}{Taghi~M
  Khoshgoftaar}.} \bibinfo{year}{2009}\natexlab{}.
\newblock \showarticletitle{A survey of collaborative filtering techniques}.
\newblock \bibinfo{journal}{\emph{Advances in Artificial Intelligence}}
  \bibinfo{volume}{2009} (\bibinfo{year}{2009}), \bibinfo{pages}{4}.
\newblock


\bibitem[\protect\citeauthoryear{Wang, Liu, Kumar, and Chang}{Wang
  et~al\mbox{.}}{2016}]%
        {wang2016learning}
\bibfield{author}{\bibinfo{person}{Jun Wang}, \bibinfo{person}{Wei Liu},
  \bibinfo{person}{Sanjiv Kumar}, {and} \bibinfo{person}{Shih-Fu Chang}.}
  \bibinfo{year}{2016}\natexlab{}.
\newblock \showarticletitle{Learning to hash for indexing big data-A survey}.
\newblock \bibinfo{journal}{\emph{Proc. IEEE}} \bibinfo{volume}{104},
  \bibinfo{number}{1} (\bibinfo{year}{2016}), \bibinfo{pages}{34--57}.
\newblock


\bibitem[\protect\citeauthoryear{Wang, Ruan, Zhang, and Si}{Wang
  et~al\mbox{.}}{2013}]%
        {wang2013learning}
\bibfield{author}{\bibinfo{person}{Qifan Wang}, \bibinfo{person}{Lingyun Ruan},
  \bibinfo{person}{Zhiwei Zhang}, {and} \bibinfo{person}{Luo Si}.}
  \bibinfo{year}{2013}\natexlab{}.
\newblock \showarticletitle{Learning compact hashing codes for efficient tag
  completion and prediction}. In \bibinfo{booktitle}{\emph{Proceedings of the
  22nd ACM International Conference on Information \& Knowledge Management}}.
  ACM, \bibinfo{pages}{1789--1794}.
\newblock


\bibitem[\protect\citeauthoryear{Weiss, Torralba, and Fergus}{Weiss
  et~al\mbox{.}}{2009}]%
        {weiss2009spectral}
\bibfield{author}{\bibinfo{person}{Yair Weiss}, \bibinfo{person}{Antonio
  Torralba}, {and} \bibinfo{person}{Rob Fergus}.}
  \bibinfo{year}{2009}\natexlab{}.
\newblock \showarticletitle{Spectral hashing}. In
  \bibinfo{booktitle}{\emph{Advances in Neural Information Processing
  Systems}}. \bibinfo{pages}{1753--1760}.
\newblock


\bibitem[\protect\citeauthoryear{Zhang, Shen, Liu, He, Luan, and Chua}{Zhang
  et~al\mbox{.}}{2016}]%
        {zhang2016discrete}
\bibfield{author}{\bibinfo{person}{Hanwang Zhang}, \bibinfo{person}{Fumin
  Shen}, \bibinfo{person}{Wei Liu}, \bibinfo{person}{Xiangnan He},
  \bibinfo{person}{Huanbo Luan}, {and} \bibinfo{person}{Tat-Seng Chua}.}
  \bibinfo{year}{2016}\natexlab{}.
\newblock \showarticletitle{Discrete collaborative filtering}. In
  \bibinfo{booktitle}{\emph{Proceedings of the 39th International ACM SIGIR
  Conference on Research and Development in Information Retrieval}},
  Vol.~\bibinfo{volume}{16}.
\newblock


\bibitem[\protect\citeauthoryear{Zhang, Wang, Ruan, and Si}{Zhang
  et~al\mbox{.}}{2014}]%
        {zhang2014preference}
\bibfield{author}{\bibinfo{person}{Zhiwei Zhang}, \bibinfo{person}{Qifan Wang},
  \bibinfo{person}{Lingyun Ruan}, {and} \bibinfo{person}{Luo Si}.}
  \bibinfo{year}{2014}\natexlab{}.
\newblock \showarticletitle{Preference preserving hashing for efficient
  recommendation}. In \bibinfo{booktitle}{\emph{Proceedings of the 37th
  International ACM SIGIR Conference on Research \& Development in Information
  Retrieval}}. ACM, \bibinfo{pages}{183--192}.
\newblock


\bibitem[\protect\citeauthoryear{Zhou, Li, Zhao, Chen, Li, Yang, Cui, Yu, Chen,
  Ding, et~al\mbox{.}}{Zhou et~al\mbox{.}}{2017}]%
        {zhou2017kunpeng}
\bibfield{author}{\bibinfo{person}{Jun Zhou}, \bibinfo{person}{Xiaolong Li},
  \bibinfo{person}{Peilin Zhao}, \bibinfo{person}{Chaochao Chen},
  \bibinfo{person}{Longfei Li}, \bibinfo{person}{Xinxing Yang},
  \bibinfo{person}{Qing Cui}, \bibinfo{person}{Jin Yu}, \bibinfo{person}{Xu
  Chen}, \bibinfo{person}{Yi Ding}, {et~al\mbox{.}}}
  \bibinfo{year}{2017}\natexlab{}.
\newblock \showarticletitle{KunPeng: Parameter Server based Distributed
  Learning Systems and Its Applications in Alibaba and Ant Financial}. In
  \bibinfo{booktitle}{\emph{Proceedings of the 23rd ACM SIGKDD International
  Conference on Knowledge Discovery and Data Mining}}. ACM,
  \bibinfo{pages}{1693--1702}.
\newblock


\bibitem[\protect\citeauthoryear{Zhou and Zha}{Zhou and Zha}{2012}]%
        {zhou2012learning}
\bibfield{author}{\bibinfo{person}{Ke Zhou} {and} \bibinfo{person}{Hongyuan
  Zha}.} \bibinfo{year}{2012}\natexlab{}.
\newblock \showarticletitle{Learning binary codes for collaborative filtering}.
  In \bibinfo{booktitle}{\emph{Proceedings of the 18th ACM SIGKDD International
  Conference on Knowledge Discovery and Data Mining}}. ACM,
  \bibinfo{pages}{498--506}.
\newblock


\end{thebibliography}

\end{document}